\documentclass{ieeeaccess}
\usepackage{graphicx}%
\usepackage{multirow}%
\usepackage{amsmath,amssymb,amsfonts}%
\usepackage{amsthm}%
\usepackage{mathrsfs}%
\usepackage{textcomp}%
\usepackage{manyfoot}%
\usepackage{booktabs}%
\usepackage{algorithm}%
\usepackage{algorithmicx}%
\usepackage{algpseudocode}%
\usepackage{listings}%
\usepackage{float}
\usepackage{booktabs}
\usepackage{subfigure}
\usepackage{caption}
\usepackage{tabularx}
\usepackage[numbers,sort,compress]{natbib}
\usepackage{pdfpages}
\usepackage{fancyhdr}
\def\BibTeX{{\rm B\kern-.05em{\sc i\kern-.025em b}\kern-.08em
    T\kern-.1667em\lower.7ex\hbox{E}\kern-.125emX}}
\begin{document}
%\chead{ First article title}
\cfoot{This article has been accepted for publication in IEEE Access. This is the author\'s version which has not been fully edited and content may change prior to final publication. Citation information: DOI $10.1109/ACCESS.2024.3428847$}
\history{Date of publication 7/16/2024. \\ The paper is accepted to be published in IEEE Access. \\ This is the accepted version of the article which has not been fully edited and content may change prior to final publication. \\The doi of the article to access the final version along with refereing to this article (citation information) is presented bellow.}
\doi{10.1109/ACCESS.2024.3428847}

\title{An Explainable Deep Learning-Based Method For Schizophrenia Diagnosis Using Generative Data-Augmentation}
\author{\uppercase{Mehrshad Saadatinia}\authorrefmark{1},
\uppercase{Armin Salimi-Badr \authorrefmark{1}, 
\IEEEmembership{Senior Member, IEEE}}
\address[1]{Faculty of Computer Science and Engineering, Shahid Beheshti University, Tehran, Iran}}

\markboth
{M. Saadatinia and A. Salimi-Badr \headeretal: An Explainable Deep Learning-Based Method For Schizophrenia Diagnosis Using Generative Data-Augmentation}
{M. Saadatinia and A. Salimi-Badr \headeretal: An Explainable Deep Learning-Based Method For Schizophrenia Diagnosis Using Generative Data-Augmentation}

\corresp{Corresponding author: Armin Salimi-Badr (e-mail: a\_salimibadr@sbu.ac.ir).}

\begin{abstract}
Schizophrenia is an example of a rare mental disorder that is challenging to diagnose using conventional methods. Deep learning methods have been extensively employed to aid in the diagnosis of schizophrenia. However, their efficacy relies heavily on data quantity, and their black-box nature raises trust concerns, especially in medical diagnosis contexts. In this study, we leverage a deep learning-based method for the automatic diagnosis of schizophrenia using EEG brain recordings. This approach utilizes generative data augmentation, a powerful technique that enhances the accuracy of the diagnosis. Additionally, our study provides a framework to use when dealing with the challenge of limited training data for the diagnosis of other potential rare mental disorders. To enable the utilization of time-frequency features, spectrograms were extracted from the raw signals. After exploring several neural network architectural setups, a proper convolutional neural network (CNN) was used for the initial diagnosis. Subsequently, using Wasserstein GAN with Gradient Penalty (WGAN-GP) and Variational Autoencoder (VAE), two different synthetic datasets were generated in order to augment the initial dataset and address the over-fitting issue. The augmented dataset using VAE achieved a 3.0\% improvement in accuracy, reaching 99.0\%, and also demonstrated faster convergence. Finally, we addressed the lack of trust in black-box models using the Local Interpretable Model-agnostic Explanations (LIME) algorithm to determine the most important superpixels (frequencies) in the diagnosis process.
\end{abstract}


\titlepgskip=-15pt

\maketitle

\section{Introduction}\label{sec1}

Schizophrenia is a severe mental illness from which around 1\% of the population suffers. Commonly a schizophrenia patient develops auditory hallucinations, unusual beliefs, and a deterioration in cognitive abilities and social function and is diagnosed with schizophrenia in early adulthood \cite{10.1001/jamapsychiatry.2019.3360}. In order to diagnose schizophrenia there exist no objective diagnostic tests. Therefore, diagnosis is based on observed behavior, a psychiatric history that includes the person's reported experiences, and reports of others familiar with the person. To be diagnosed with schizophrenia, symptoms must be present over an extended period, typically six months \cite{guha2014diagnostic}. 

The use of computer-aided methods for the diagnosis of such illnesses could significantly help experts in the process of diagnosis since they are typically accurate and fast compared to current methods of diagnosis. This can potentially mitigate or fully prevent the permanent damage to the brain caused by a late diagnosis of schizophrenia. Machine learning methods have been by far the most popular computer-aided approach for medical applications including the diagnosis of such diseases in recent years with growing popularity \cite{bagherzadeh2022detection,salimi2020neural, salimi2023early,nabavi2021medical,salimi2022type,ACC1, ACC2, ACC3,ACC4}. 

A major drawback of traditional supervised machine learning methods for medical diagnosis is that they require a level of domain expertise in order to find and extract useful features to be fed to the model. However, this issue could be addressed using deep learning methods through representation learning. 

Generally, deep learning methods work best with very large datasets because they need to extract features and find patterns in the data without any prior feature extraction or assistance. Using such methods for medical diagnosis poses challenges for researchers, as data collection for these studies requires specialized facilities and finding subjects. These challenges are more significant for rare diseases and disorders like schizophrenia. Deep learning-based diagnosis of rare mental disorders, especially schizophrenia, has been extensively explored in past studies. Some of these studies have employed innovative approaches and novel network architectures to enhance the classification accuracy of the models. However, accuracy improvements achieved through enhancing network architectures are limited in the absence of sufficient data samples. The most crucial factor for training robust deep learning models is the availability of ample training data.

Lack of sufficient training data is a tremendous challenge for medical applications, and more specifically for mental disorders diagnosis using deep learning. There are several techniques to deal with the lack of data and still achieve good results \cite{alzubaidi2023survey}. Transfer learning and data augmentation are most popular methods commonly used for this purpose. Transfer learning has been effective in many instances \cite{shalbaf2020transfer, zheng2021diagnosis}. However, it still relies on the assumption that there is sufficient training data for another similar task to train a robust model. Data augmentation is the other popular method. To augment the datasets at hand there are several approaches ranging from traditional basic techniques to more novel deep learning-based methods \cite{chlap2021review}. Many studies use traditional approaches which involve manipulations and operations on the available dataset instances. However these approaches are not always suitable since manipulating the data might effect the integrity of the data for some modalities, EEG being one of them. Generating synthetic data using deep generative models to address this challenge has been extensively used in various domains across different studies. In the context of medical diagnosis, particularly for mental disorders, MRI and fMRI data are the most commonly used data formats. Consequently, most studies have employed deep generative models to generate synthetic data for MRI and fMRI modalities \cite{liu2022attention, 8759585}. EEG data is another popular modality used for the analysis of brain activity. However, there is clearly a gap in the related studies concerning the application of deep generative models to generate high-fidelity synthetic EEG data. While some studies have aimed at using generative approaches for EEG data generation \cite{habashi2023generative}, the field could still benefit from innovations and the proposal of different approaches and architectures in this regard.

Moreover, one issue in using deep neural networks for medical applications is their black-box structure \cite{gunning2019xai}. Indeed, it is not clear why the model decides to categorize a person as a patient or a healthy one which is required to make these systems trustworthy \cite{tjoa2020survey,salimi2022type}. Countless studies have used deep learning models for the diagnosis of various diseases and disorders, but very few have aimed to explain their models' internal workings using explainable algorithms capable of providing insight into these non-linear deep models. Consequently, to increase the explainability of our method and address this gap, we use the LIME algorithm to detect the features in the input that contribute to a positive or negative diagnosis. This approach effectively mitigates the issue of lack of trustworthiness in medical diagnosis systems for our study.

The remainder of this paper is structured as follows: First we review the related studies and present the contributions of the current article. Next, Section II describes the fundamental concepts and preliminaries necessary for this study and  Section III the methodology used in this study. Section IV discusses the experiments and the results of each and finally in the last section we conclude our work.

\subsection{Related Works} 
Several researchers have previously proposed different deep learning and machine learning-based methods for schizophrenia diagnosis using Electroencephalography (EEG) signals. The literature is quite rich in schizophrenia diagnosis using traditional machine learning. However, we only focus on deep learning-based methods in this study. 
Related works on deep learning-based schizophrenia diagnosis using EEG signals are summarized briefly as follows.

So far many studies have focused on utilizing deep neural networks for schizophrenia diagnosis using EEG signals \cite{Verma_2023}. In the work of Chu et al. \cite{chu2017individual} new systems are proposed for individual recognition based on spatiotemporal resting state EEG data, using modified deep learning architectures. The proposed system achieved high degrees of classification accuracy, with 81.6\% accuracy for high-risk individuals, 96.7\% for clinically stable first-episode patients with schizophrenia, and 99.2\% for healthy controls. The study also found that replacing the softmax layer with Random Forest and adding a voting layer improved the classification accuracy.
The choice of Convolutional Neural Network (CNN) as the type of neural network has been the most widely-adopted approach in the relevant literature. However, the past studies mainly differ in terms of the choice of data representation as neural network inputs.
In the study of Naira et al. \cite{Naira2019} the Pearson Correlation Coefficient (PCC) matrix between channels has been used as input to a Convolutional Neural Network (CNN), and a 90\% accuracy was obtained. Another study \cite{bagherzadeh2022detection} used Transfer Entropy (TE) values between channels to represent the signals as 2D matrices and then classify them using a CNN-LSTM architecture, obtaining a near-perfect accuracy over a 19-channel dataset.
Oh et al. \cite{app9142870} used 1D-CNN and the proposed model achieved classification accuracy of 98.07\% and 81.26\% for non-subject-based testing and subject-based testing, respectively. using a hybrid 1D-CNN and LSTM architecture has also been explored by Shoeibi et al. \cite{shoeibi2021automatic}. 

The most impressive results have been achieved by the studies which have taken time and frequency features of EEG recordings into account. Phang et al. \cite{8836535} proposed a novel approach, using brain functional connectivity in order to discriminate healthy subjects from schizophrenic subjects. Their method estimates time-frequency connectivity measures as well as topological-based complex network measures, using separate 2D and 1D CNNs to learn latent representations. Their method achieved an accuracy of 91.69\% on the 16-channel EEG dataset which is also used in the present study. Sun et al. \cite{sun2021hybrid} proposed a hybrid neural network architecture of CNN and LSTM. They used Fast Fourier Transform (FFT) and Fuzzy Entropy (FuzzyEn) algorithms for frequency and time domain feature extraction respectively and compared the classification result of each one using their hybrid architecture. They were able to achieve an accuracy of 99.22\% using FuzzyEn time-domain features. Several studies in the past few years have utilized different methods for incorporating Short-Time Fourier Transform (STFT) and Continuous Wavelet Transform (CWT) into the EEG-based diagnosis of schizophrenia using deep learning \cite{sahu2023scz, shalbaf2020transfer, khare2021spwvd, aslan2020automatic}. We have presented an extensive comparison and analysis of these studies in the subsequent sections of this study.

The related studies clearly demonstrate that the choice of neural network architecture and input representation significantly impacts the effectiveness of deep learning-based methods for schizophrenia diagnosis using EEG signals. However, the results obtained in prior studies on deep learning-based schizophrenia diagnosis mostly stem from experimenting with diverse CNN architectures and data representations. 

In the work of Shalbaf et al. \cite{shalbaf2020transfer} the effectiveness of transfer learning for schizophrenia diagnosis has been explored. Sobahi et al. have explored limited data augmentation using Extreme learning machines (ELM) based autoencoders and combined with transfer learning \cite{sobahi2022new}. Their approach achieved a 97.7\% accuracy which stems from both data augmentation they used and transfer learning.

Although some studies have utilized generative data augmentation with MRI and fMRI datasets, in our study we aim to explore innovative generative methods specifically tailored for generating high-fidelity synthetic EEG samples for diagnosis of mental disorders like schizophrenia to potentially build upon and further improve the previous approaches. 

\subsection{Our Contributions} 
Based on the considerations above, Our approach uses spectrograms as the time-frequency representation of the EEG signals and utilizes CNN for the classification of healthy and schizophrenic subjects.  Importantly, to the best of our knowledge, this study represents the first instance of incorporating generative data augmentation solely to enhance deep learning-based schizophrenia diagnosis.
Our main contributions are summarized as follows:
\begin{enumerate}
\item Introducing an instantiation of CNN architecture, inspired by VGGNet-like architectures and specifically crafted for our problem domain and computational resources. Our contribution in this regard lies in the strategic arrangement and configuration of existing layers to exhibit far more robustness than well-known CNN architectures for our problem, while achieving a remarkable reduction in the number of parameters;
\item Utilizing generative models for synthetic data generation in order to augment the available dataset and improve the classification and diagnosis accuracy; 
\item Proposing specialized architectural configurations for VAE and WGAN-GP to support rectangular (non-squar) inputs and produce high-fidelity synthetic instances;
\item Using the Local Interpretable Model-agnostic Explanations (LIME) algorithm in order to detect important features in the input spectrograms for a negative or positive diagnosis.
\end{enumerate}

\section{Preliminaries}

\subsection{Data Augmentation Using Generative Models} 
This work uses spectrograms, 2D representations of time and frequency, necessitating image-based data augmentation methods. Traditional methods such as rotation, cropping, flipping, and gamma correction are effective for images \cite{wang2018alcoholism}, but can distort spectrogram data. Instead, synthetic spectrograms can be generated using generative models like Variational Autoencoders (VAEs) \cite{kingma2013autoencoding} and Generative Adversarial Networks (GANs) \cite{goodfellow2014generative}. Both models were employed and compared in this study to find the optimal approach.

VAEs, generative autoencoders, extract latent representations and encourage them to follow a target distribution using Kullback-Leibler (KL) divergence added to the reconstruction loss:

\begin{equation}
\label{eq:vaeloss1}
\mathcal{L} = \text{Reconstruction Loss} + \text{KL Divergence}
\end{equation}

%If the decoder part consists of $\theta$ as its weights, and the encoder part includes $\phi$ as its parameters, the mentioned loss in Eq. (\ref{eq:vaeloss1}) would be rewritten as follows \cite{goodfellow2016deep}:
%\begin{equation}
%\begin{array}{ll}
%     \mathcal{L}(\phi,\theta)&=\sum_{i=1}^N\mathbb{E}_{z_i \sim q_\phi(z_i|x_i)}\left(logp_\theta(x_i|z_i)\right)  \\
%     & -D_{KL}(q_\phi(z_i|x_i) || p(z_i))
%\end{array}
%\end{equation}
%where, $z_i$ is the encoder's part output for $i^{th}$ training sample, $N$ is the number of training samples, $p_\theta$ is the decoder function, $q_\phi$ is the encoder function, and $x_i$ is the input sample.

GANs consist of a generator and a discriminator competing in a minimax game, improving each other over time. The generator tries to produce realistic samples, while the discriminator distinguishes real from fake samples. Variations like Deep Convolutional GAN (DCGAN) address training instabilities with convolutional neural networks but still face issues like mode collapse \cite{radford2015unsupervised}.

\subsection{Wasserstein GAN with Gradient Penalty}
Wasserstein GAN (WGAN) modifies the GAN loss using the Wasserstein distance to align generated and real distributions \cite{arjovsky2017wasserstein}. The minimax game between the generator \( G \) and discriminator \( D \) is defined by:

\begin{equation} \label{eq:wloss}
L = \min_{G}\max_{D} \left( \mathbb{E}_{x \sim p_r(x)}[D(x)] - \mathbb{E}_{z \sim p_z(z)}[D(G(z))] \right)
\end{equation}
Here, \( G \) is the generator network that creates data samples from random noise \( z \), and \( D \) is the discriminator network that differentiates between real data \( x \) and generated data \( G(z) \). The term \( p_r(x) \) represents the true data distribution, and \( p_z(z) \) represents the distribution of the noise. Indeed, the main aim of the GAN is to generate a sample $G(z)$ similar to real data $x$ receiving the input noise $z$.

During training, the generator and discriminator are trained iteratively, with the discriminator trained several times per generator iteration. A crucial constraint is that the discriminator must be 1-Lipschitz continuous, ensuring bounded gradients. This is enforced by WGAN-GP, which adds a gradient penalty term to the loss function \cite{gulrajani2017improved}:

\begin{equation} \label{eq:wgp}
\lambda \mathbb{E}_{\hat{x} \sim p_{\hat{x}}(\hat{x})}\left[\left(\|\nabla_{\hat{x}} D(\hat{x})\|_2 - 1\right)^2\right]
\end{equation}

In this equation, \( \lambda \) is the regularization parameter, \( \hat{x} \) represents interpolated data samples, and \( p_{\hat{x}}(\hat{x}) \) denotes the distribution of these interpolated data samples.

WGAN-GP offers more stable training, an intuitive loss value, and higher quality outputs. For these reasons, WGAN-GP and VAE were used for synthetic data generation in this study, with VAEs favored for their lower training cost and faster results.

\subsection{LIME Algorithm}
The Local Interpretable Model-agnostic Explanations (LIME) algorithm, is designed to enhance the explainability of complex, black-box machine learning models (such as the CNN) by providing local and simplified explanations for individual predictions \cite{ribeiro2016should}. To employ the LIME algorithm, We start by selecting a specific data point from the dataset that we are interested in an explanation for. Perturbations are then generated by randomly altering the features of the chosen data point, representing variations around the original instance. Subsequently, predictions for these perturbed instances are obtained using the black-box model. A local interpretable model, such as linear regression, is fitted to the perturbed instances, treating the model's predictions as the target variable and the perturbed instances as input features. The instances are subjected to weighted sampling based on their proximity to the original data point, and the local model is trained on this weighted sample. The coefficients of the local model are then utilized as explanations for the predictions made by the original black-box model at the initially selected data point. This process helps reveal insights into the model's decision-making for that specific instance in an explainable manner.

\section{Methodology}

\begin{figure*}[!t]
    \centering
    \includegraphics[width=7.0in]{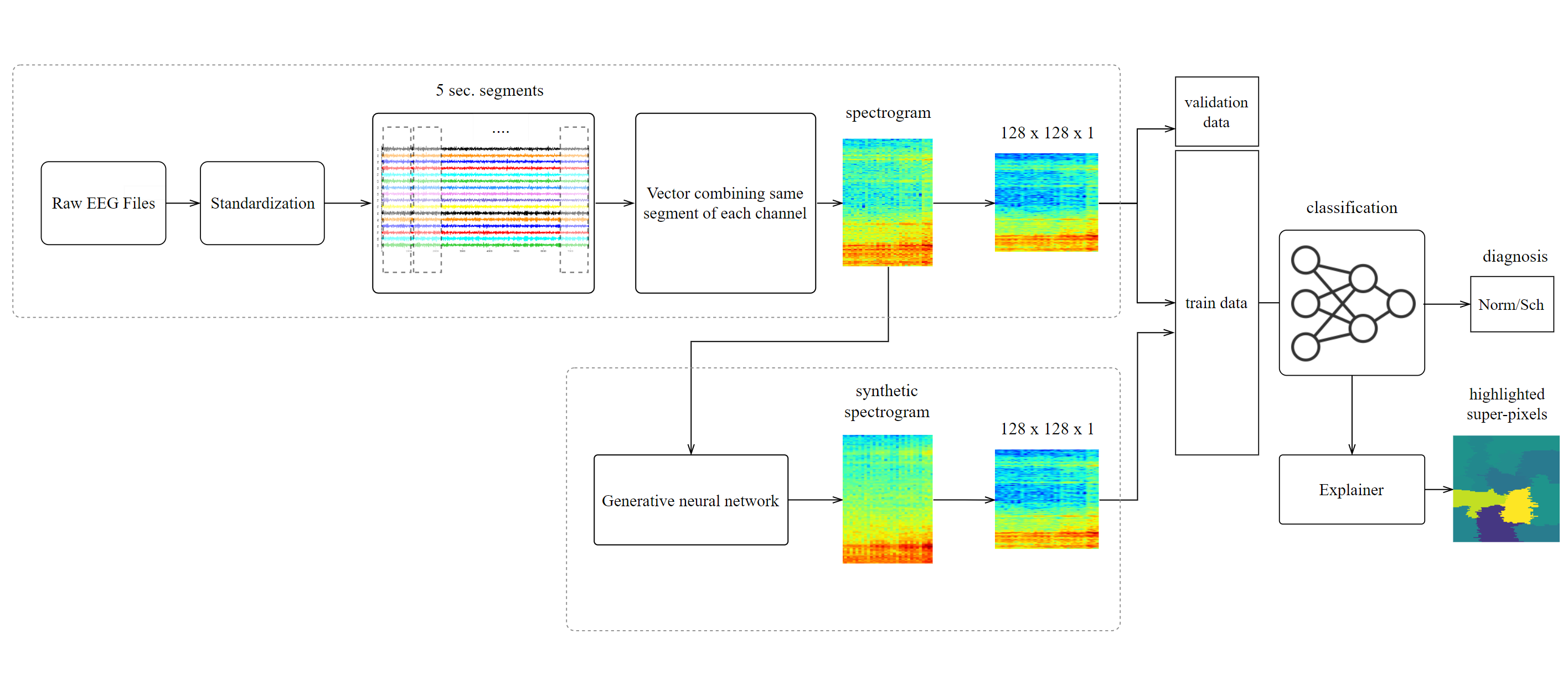}
    \caption{An overview of the methodology used for diagnosis and explanation. The bottom branch represents the synthetic data generations}
    \label{fig:methoddiag}
\end{figure*}

We propose a deep learning-based approach for the diagnosis of schizophrenia using generative data augmentation and explainable algorithms. Our methodology involves several key steps. In this section, all the steps are described. Some of these steps involve conducting several experiments. The details of these experiments, including parameters and results, are provided in the subsequent section. Figure \ref{fig:methoddiag} illustrates an overview of the methods and stages of this study including diagnosis, data augmentation, and the explanations of the model, all of which are discussed in further detail throughout the rest of the paper. 

\subsection{Datasets} 
Two datasets were used in this study. The details of each are described below.
\begin{itemize}
\item 
The first and main dataset consists of 16-channel EEG brain recordings\footnote{Available at: http://brain.bio.msu.ru/eeg\_schizophrenia.htm.}. This dataset will be referred to as the "first dataset" or "16-channel" throughout the rest of this document. The recordings belong to two groups. The first group consists of 39 healthy adolescents (11-13 years old). The second consists of 45 adolescents (10-14 years old) diagnosed with schizophrenia. These EEG recordings were recorded with a sampling frequency of 128 Hz, while, subjects were in a closed-eye resting state. The recording length is 60 seconds and the recorded channels are F7, F3, F4, F8, T3, C3, Cz, C4, T4, T5, P3, Pz, P4, T6, O1, O2. Each subject in this dataset is represented in a text file, each consisting of all the samples for consecutive channels in a single column. After reading and separating the channels, we have 16x7680 matrices for each subject. 

\item The second dataset has 19 channels. As described by the authors of this dataset "it comprised 14 patients with paranoid schizophrenia and 14 healthy controls. Data were acquired with the sampling frequency of 250 Hz using the standard 10-20 EEG montage with 19 EEG channels: Fp1, Fp2, F7, F3, Fz, F4, F8, T3, C3, Cz, C4, T4, T5, P3, Pz, P4, T6, O1, O2. The reference electrode was placed between electrodes Fz and Cz" \cite{repod.0107441_2017}. This dataset is stored in standard .edf format and after reading the data samples, the same processing steps as the first dataset have been used for it. However, not all recordings in this dataset have an equal number of samples. Therefore, the recordings for all subjects have been cut to 185000 samples which is the least across both classes. The remaining samples were thrown away and for each subject, 19x185000 matrices were extracted. 
\end{itemize}

\begin{figure}
  \centering
 \begin{tabular}{c}
    \subfigure[][]{\includegraphics[width = 3.5in]{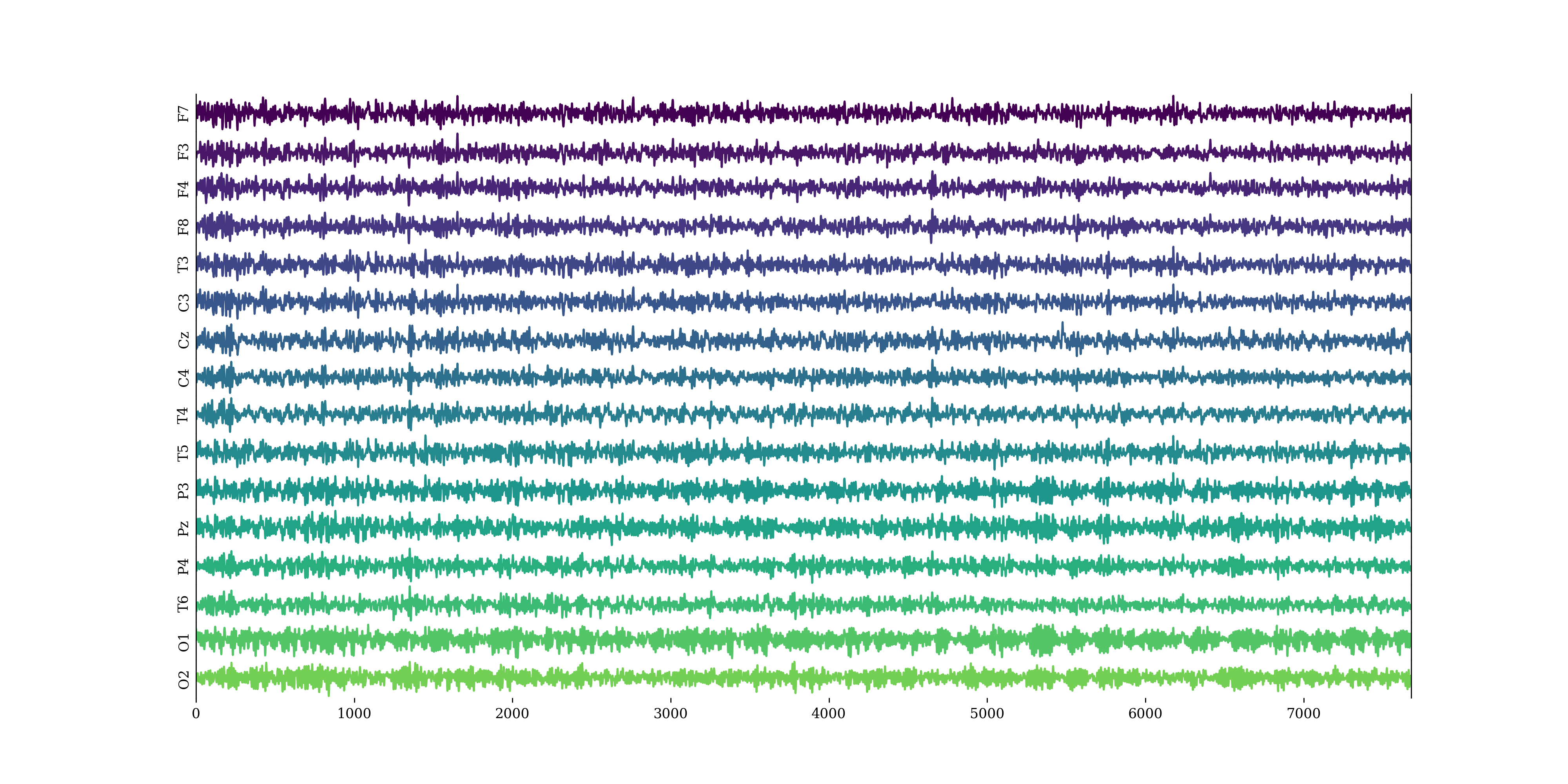}} \\
    \subfigure[][]{\includegraphics[width = 3.5in]{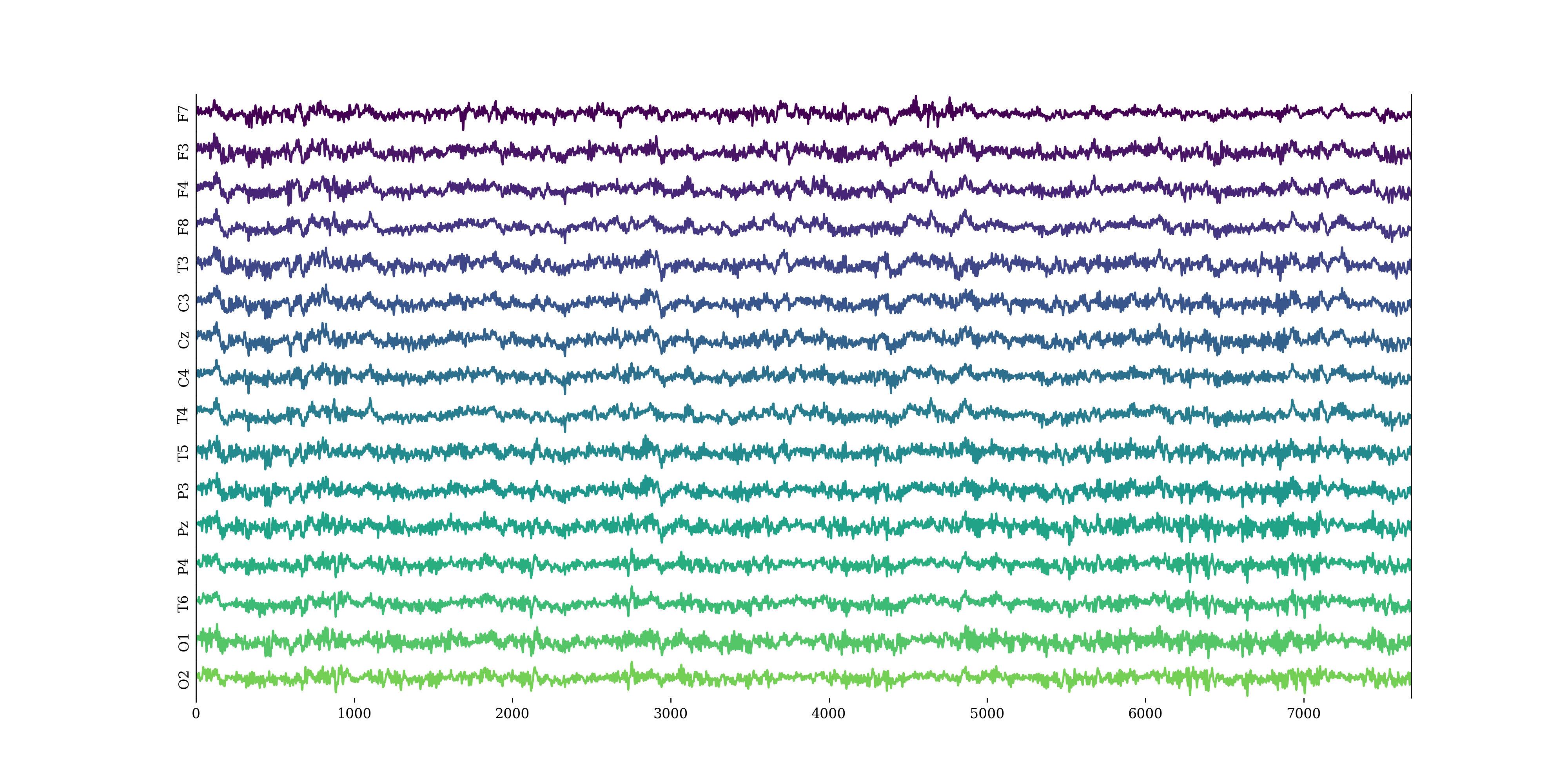}} 
\end{tabular}
  \caption{A 16-channel EEG signal with 7680 samples of (a) healthy subject, and (b) schizophrenic patient}
  \label{fig:16eegs-main}
\end{figure}

\begin{figure}
  \centering
  \begin{tabular}{c}
    \subfigure[][]{\includegraphics[width = 3.5in]{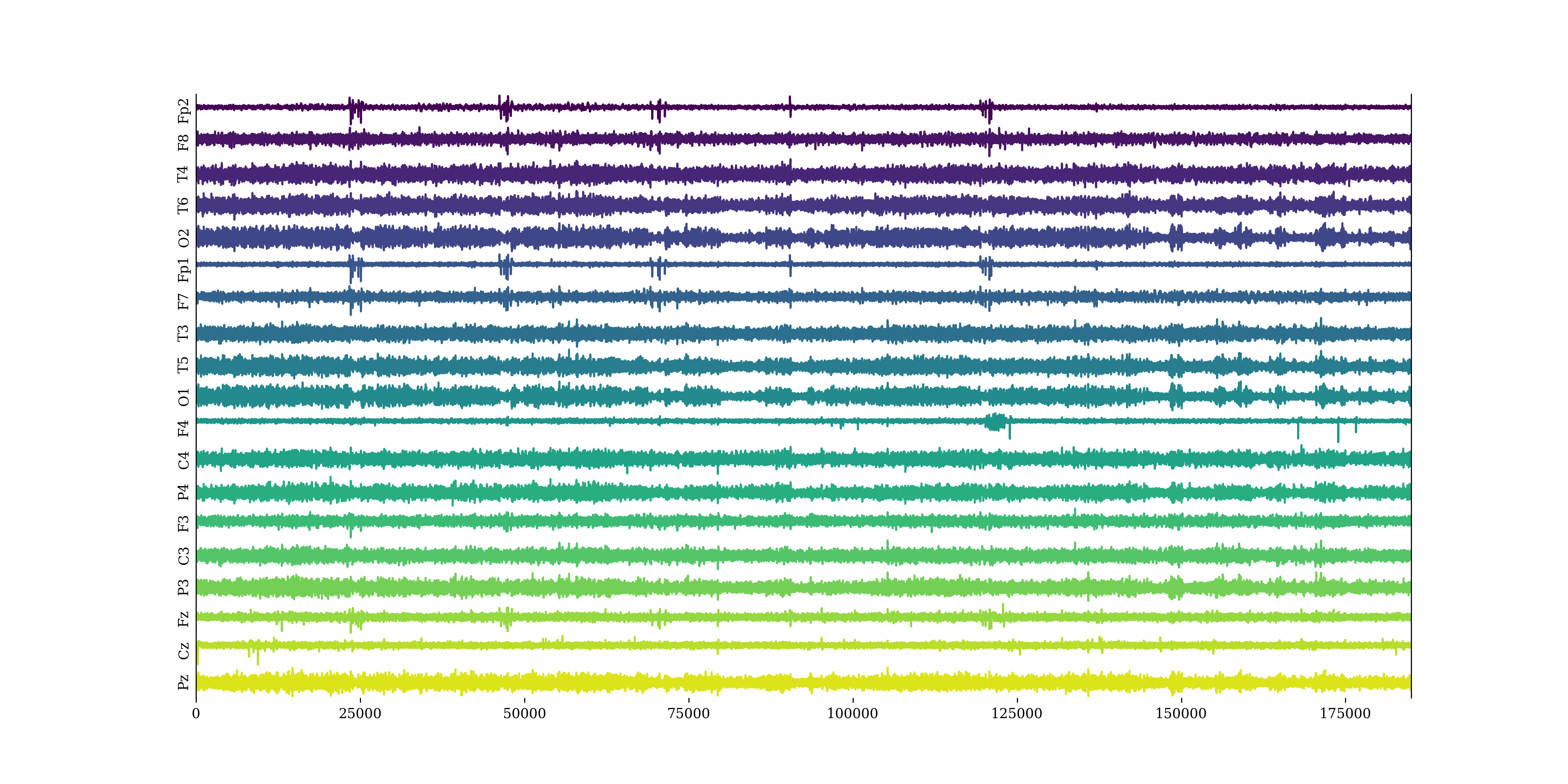}} \\
    \subfigure[][]{\includegraphics[width = 3.5in]{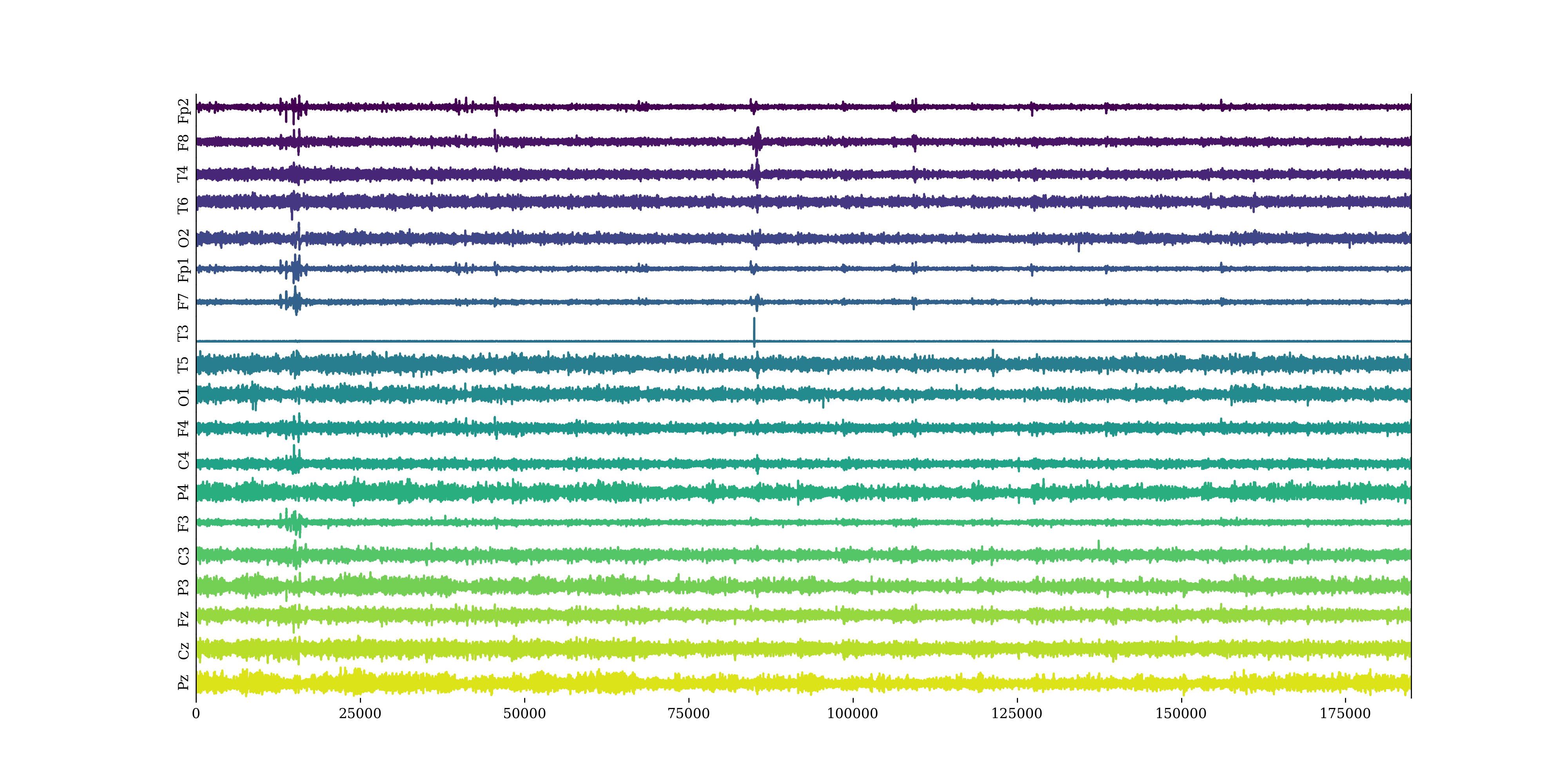}} 
\end{tabular}
  \caption{A 19-channel EEG signal with 185000 samples of (a) healthy subject, and (b) schizophrenic patient}
  \label{fig:19eegs-main}
\end{figure}

\subsection{EEG Data Processing}
\begin{figure}[t]
    \centering
    \begin{tabular}{c}
    \subfigure[]{\includegraphics[width=3.5in]{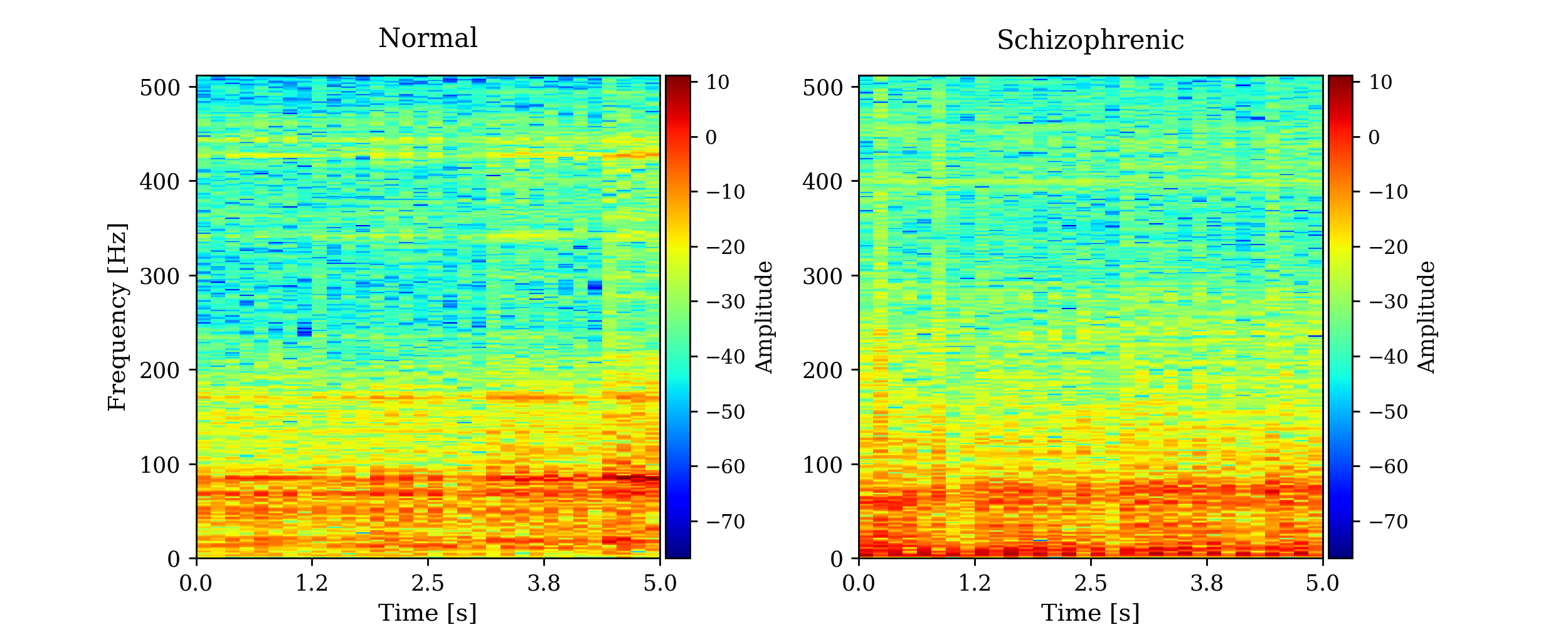}}\\
    \subfigure[]{\includegraphics[width=3.5in]{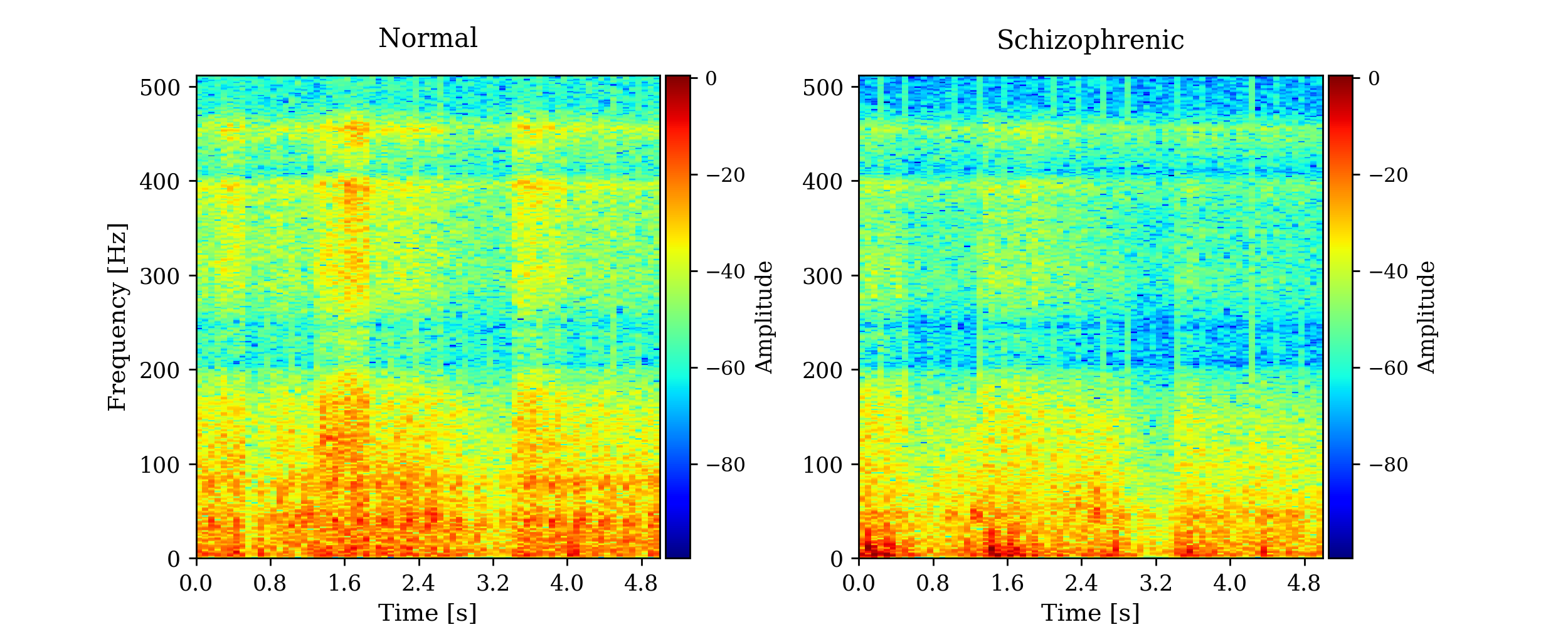}}
    \end{tabular}
    \caption{The spectrograms of 5-second segments obtained from: (a) 16-channel dataset, and (b) 19-channel dataset.}
    \label{fig:specA}
\end{figure}

After reading the EEG signals for each subject, the channels are initially separated and standardized using the z-score normalization method to bring the EEG values (voltages) across different subjects within the same range. 
In order to increase the dataset size signals are divided into equal-sized smaller segments. The choice of segment length involves a trade-off. Opting for a smaller segment size creates more training samples, but using overly-small segments may misrepresent the data and hinder the model's ability to correctly learn features. The related past studies have mostly selected shorter segment sizes like 4, 5, or 6 seconds \cite{sahu2023scz, aslan2020automatic, sun2021hybrid}. While some other studies have opted for longer segments like 25 seconds \cite{shoeibi2021automatic}. In this study, we have selected 5-second long segment sizes. Therefore, each channel is divided into 5-second segments, and the segments from every channel are combined into a single vector. Each of these vectors represents the combined channels for every 5-second segment. These vectors are used to generate spectrograms, which are 2D representations of time against the frequency and are generated using the Short-Time Fourier Transform (STFT). The spectrograms were extracted using the SciPy signal processing library in Python programming language. The number of Fast Fourier Transforms (NFFT) parameter was selected as 1022 so that the output has 512 frequency segments. The NPERSEG parameter which represents the number of points per each time segment in the spectrogram was also set to 360 for both datasets. This creates 512x32 spectrograms for the 16-channel dataset and 512x75 for the 19-channel dataset. The 16-channel dataset yields 12 segments for each of the 84 subjects, resulting in a total dataset size of 1008 spectrograms. In contrast, the 19-channel dataset generates significantly more instances due to its longer duration of 740 seconds. As a result, the second dataset yields a total of 4144 spectrograms. Since these spectrograms do not have equal dimensions across both datasets, they have been resized to 128x128 and then stored as arrays. This makes the dimensions of the data equal across the two datasets, hence, simplifying and generalizing the CNN network architecture used for classification. However, for artificial data synthesis, the original spectrograms were used. Figure \ref{fig:specA} illustrates the original spectrograms of 5-second segments for 16-channel and 19-channel datasets prior to resizing.


We initially processed the Raw EEG File, performed the necessary operations such as standardization, and then carried out the segmentation of the signals. We obtained spectrograms of our segmented EEG signals and resized these spectrograms to 128x128 to make the subsequent CNN architecture simpler.

\subsection{Classification Using Real Spectrograms}
\begin{figure}[t]
    \centering
    \includegraphics[width=0.9\linewidth]{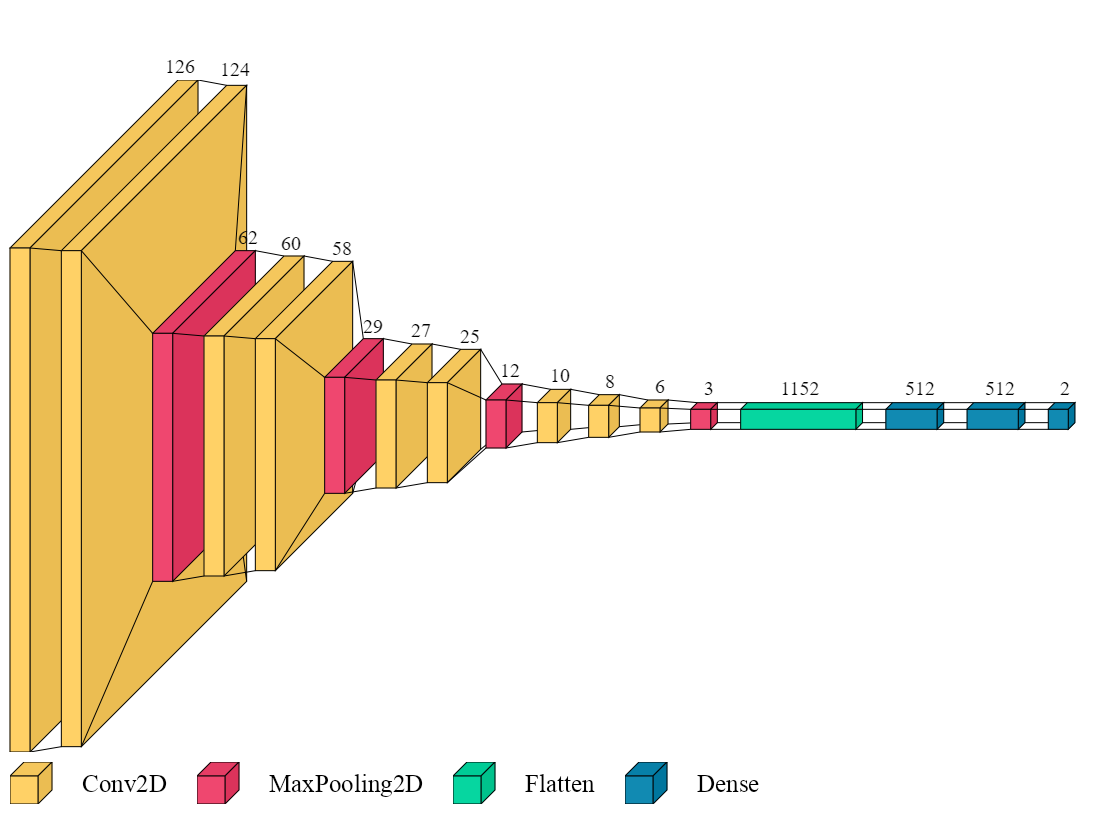}
    \caption{The proposed CNN architecture}
    \label{fig:architecture}
\end{figure}

In this stage of the study, the resized 128x128 spectrograms were fed to our proposed CNN and the initial classification was carried out. For this purpose, we proposed a CNN architecture that receives 128x128 resized spectrograms as inputs and contains nearly 1.3 million parameters. The proposed architecture is shown in Figure \ref{fig:architecture}. This CNN was based on VGGNet-like architectures but was made smaller and less complex to fit our use case. The design procedure followed established guidelines for designing CNN architectures commonly used in other studies and was refined through targeted experiments. 

\subsection{Generating Synthetic Spectrograms}
To generate more training data and help the model generalize better on the 16-channel dataset, two different generative models were used: WGAN-GP and VAE. Both the VAE and WGAN models utilized the original spectrograms (not resized) with dimensions of 512x32, as resizing operations involve interpolation, leading to an estimate rather than an exact representation of the original spectrogram. By using the unaltered spectrograms, we can synthesize samples that closely resemble the originals.While it is possible to use the interpolated resized spectrograms, interpolation adds additional complexity to the learning process for the generative network. Using the original spectrograms leads to faster and more effective learning. Moreover, using resized spectrograms would teach the network to generate interpolated spectrograms, making it impossible to reconstruct the actual EEG signals from these spectrograms since interpolation is an irreversible process. Nonetheless, these original spectrograms present additional challenges for designing the generative networks due to their rectangular shapes.

Generative networks were trained and subsequently used to generate synthetic data samples to augment the original dataset. Both VAE and WGAN were employed, and the results were compared. Augmenting the original dataset has two benefits: first, it can reduce overfitting when the model reaches its learning capacity, even with architectural changes or regularization methods. Additionally, it can mitigate data imbalance, which is present in our case. Synthetic samples were only added to the training dataset since adding synthetic data to the testing dataset can lead to falsely increased accuracy and inaccurate results.

Tables \ref{tab:disc} and, \ref{tab:gen} show the details of the network design for the discriminator and generator of the Wasserstein GAN with gradient penalty (WGAN-GP). Instead of using a conditional GAN in this study, separate GANs were trained for generating synthetic instances of each class.

\begin{figure}[]
    \centering
    \includegraphics[width=\linewidth]{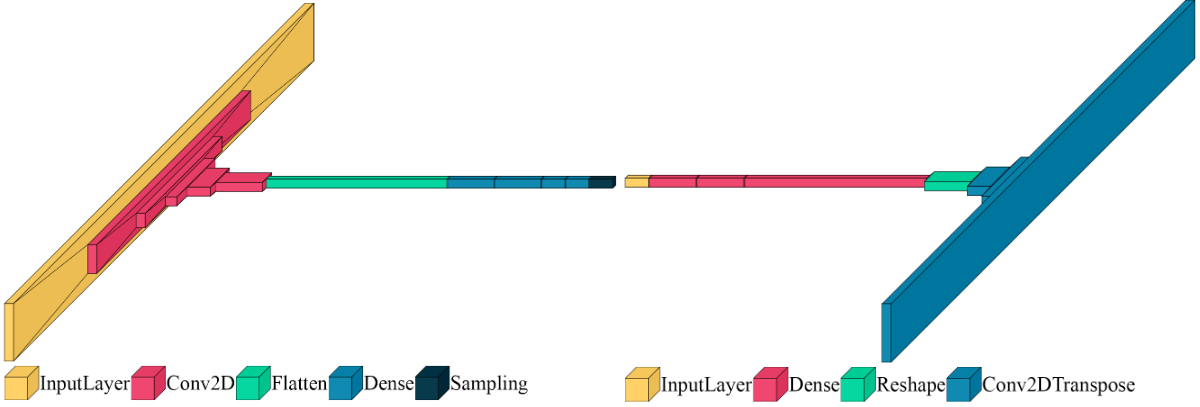}
    \caption{The proposed VAE architecture's encoder and decoder}
    \label{fig:vae}
\end{figure}

Variational autoencoder (VAE) is the other generative approach we have used in this study. Table \ref{tab:v-ae} and Figure \ref{fig:vae} illustrate the network design used for VAE. One advantage of VAE over WGAN is that it requires significantly less computational resources and every iteration takes a significantly shorter amount of time to complete which allows for longer training. Similar to the WGAN we used two separate VAEs to generate samples from each class and did not use the conditional version, an approach that in our case is feasible since we only have two classes. 

The augmentation and augmented classification process is detailed in Algorithm~\ref{alg:augmentation}.

These generated synthetic spectrograms represent 5-second EEG signals similar to the input data and since we have used the unresized spectrograms for training the generative networks, we can reconstruct these 5-second EEG segments from the synthetic spectrograms if necessary. However in our case converting the spectrograms back to EEG segments is not required.

\begin{table}
\centering
\caption{The proposed architecture for the WGAN discriminator. The activation function, dropout, and padding type are "leaky-relu (0.2)", 0.3, and "same" respectively for all layers except the output layer. The activation function of the output layer is "linear".}
\label{tab:disc}
\footnotesize
\begin{tabular}{lccc}
\toprule
{Layer}  & {kernels} & {kernel size} & {stride} \\
{}    & {(or units)} & {} & {} \\
\midrule
convolution &  32 & (5, 5) & (2, 2)\\
convolution &  64 & (5, 5) & (2, 2) \\
convolution &  128 & (5, 5) & (2, 2) \\
convolution &  256 & (5, 5) & (2, 2) \\
convolution &  512 & (5, 5) & (2, 2) \\
flatten &  8192  & - & - \\
fully-connected &  1 & - & - \\
\bottomrule
\end{tabular}
\end{table}

\begin{table}
\centering
\caption{The proposed architecture for the WGAN generator. The activation function, dropout, and padding type are "leaky-relu (0.2)", and "same" respectively for all layers except the output layer. The activation function of the output layer is "tanh" with 0.3 dropout rate. Batch normalization is applied after each "upsampling conv." layer.}
\label{tab:gen}
\footnotesize
\begin{tabular}{lccc}
\toprule
{Layer}  & {kernels} & {kernel size} & {stride} \\
{}    & {(or units)} & {} & {} \\
\midrule
input noise& 128  & - & - \\
fully-connected& 8192  & - & - \\
upsampling conv. &  256 & (3, 3) & (2, 2) \\
upsampling conv. &  128 & (3, 3) & (2, 2) \\
upsampling conv. &  64 & (3, 3) & (2, 2) \\
upsampling conv. &  32 & (3, 3) & (2, 2) \\
upsampling conv. &  1 & (3, 3) & (2, 2) \\
\bottomrule
\end{tabular}
\end{table}

\begin{table}
\centering
\caption{The proposed architecture for the variational autoencoder (VAE), (z) represents the latent vector. Padding, kernel size, and stride are "same", (5,5), and (2,2) respectively for all layers except the output layer. The kernel size of the output layer is (3,3).}
\label{tab:v-ae}
\footnotesize
\begin{tabular}{lcc}
\toprule
{Layer}  & {kernels} & {activation}\\
{}    & {(or units)} & {} \\
\midrule
convolution & 64   & relu \\
convolution & 128  & relu \\
convolution & 256  & relu\\
convolution & 512  & relu\\
convolution & 1024  & relu \\
flatten & 16384  & - \\
fully-connected & 1024  & relu \\
fully-connected & 1024  & relu \\
fully-connected (z) & 512  &linear \\
fully-connected & 1024  &relu\\
fully-connected & 1024  & relu \\
fully-connected & 16384  & relu \\
convolution transpose & 512  & relu \\
convolution transpose & 256 & relu \\
convolution transpose & 128  & relu  \\
convolution transpose & 64  & relu \\
convolution transpose & 1  & sigmoid \\
\bottomrule
\end{tabular}
\end{table}

\begin{figure}[]
    \centering
    \includegraphics[width=\linewidth]{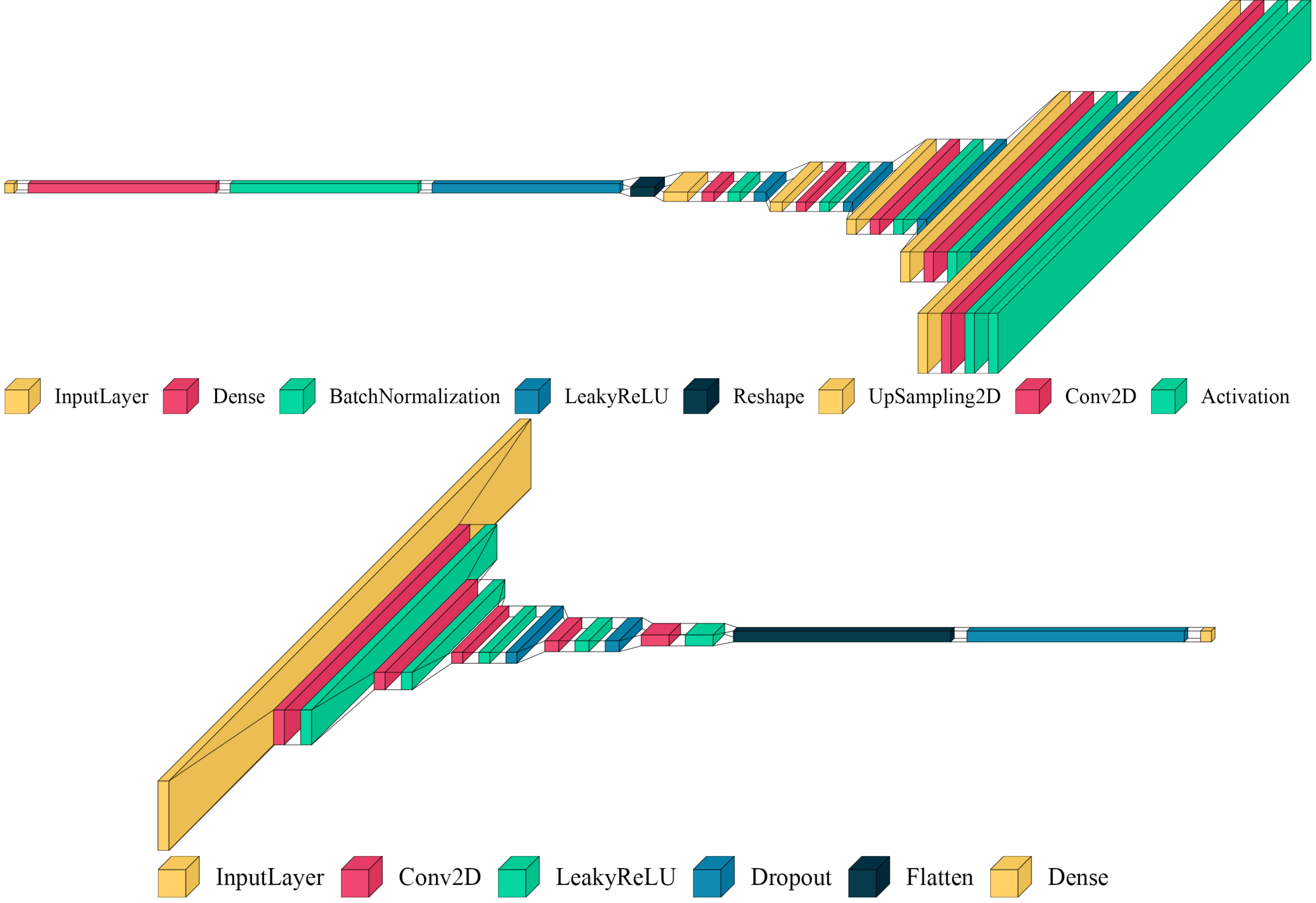}
    \caption{The proposed WGAN architecture's generator and discriminator}
    \label{fig:wgan}
\end{figure}

\begin{algorithm}
\raggedright
\small
\caption{Augmented classification with generative network (VAE or WGAN-GP)}
\label{alg:augmentation}
\textbf{Require:} spectrogram instances ($S$), generative network type ($network$),\\ number of synthetic samples for schizophrenia ($p$) and healthy ($n$), resize width ($w$) and height ($h$) for spectrograms \\
\textbf{Require:} trained generative model ($\Gamma$), trained classifier ($C$)
\\
\begin{algorithmic}[1]
\Procedure{TrainGenerativeNetwork}{$S, network$}
    \If{$network \text{ is VAE}$}
        \State $encoder, decoder \gets \Call{TrainNetwork}{S}$
        \State $\Gamma \gets decoder$
    \ElsIf{$network \text{ is WGAN-GP}$}
        \State $generator, discriminator \gets \Call{TrainNetwork}{S}$
        \State $\Gamma \gets generator$
    
    \EndIf
    \State \Return $\Gamma$
\EndProcedure
\Statex
\Procedure{AugmentedClassification}{$S, \Gamma, p, n, w, h$}
    \State $S_{\text{resized}} = \text{resize}(S, (w, h))$
    \State $train, test \gets \Call{SplitTrainTest}{S_{\text{resized}}}$
    \State $synthetic \gets \Gamma(p, n)$ 
    \State $synthetic \gets \text{resize}(synthetic, (w, h))$
    \State $train \gets train + synthetic$
    \State $C \gets \Call{TrainClassifier}{train}$
    \State \Call{Evaluate}{$C$, $test$}
    \State \Return $C$
\EndProcedure
\end{algorithmic}
\end{algorithm}

\subsection{Explanations by LIME}
Subsequently, we utilized the LIME algorithm to explain the classifier's internal classification process. Deep CNN models are black-box models, meaning that their internal working and decision-making (classification in this case) process is not clear to the user. In a medical diagnosis system, it is essential to understand what features in the input data have contributed to a specific diagnosis.  Using explainablity in the context of medical diagnosis can have several benefits:
\begin{itemize}
    \item Patients and healthcare providers are more likely to trust a diagnosis if they can understand how it was reached.
    
    \item Explainability allows for the identification and correction of potential errors in the model's decision-making process.
    
    \item Researchers and medical experts can use insights from explainable models to discover new medical knowledge, potentially leading to the identification of new indicators for diagnosing diseases or disorders.
\end{itemize}

In this study, we have used the LIME algorithm which is a model-agnostic explainer for machine learning models \cite{ribeiro2016should}. LIME works by creating perturbations of input samples and inspecting the corresponding output for each perturbation. This can help us understand what features impact the output and explain the model this way. For certain recognition tasks, such as classifying intuitive objects like animals and buildings, we can verify if the model is functioning as expected by examining the explainer’s output. In our study, however, we deal with spectrograms, which are not intuitive objects, making standalone explanations less helpful. By analyzing multiple explanation outputs however, we can identify consistent patterns for the same classes and figure out which frequency levels are of more significance for diagnosis of each class.

\section{Experiments and Results}

%\begin{figure}[H]
%    \centering
%    \includegraphics[width=0.8\linewidth]{figures/specs.eps}
%    \caption{Sample spectrograms used for training, after resizing to 224x224}
%    \label{fig:specsubplot}
%\end{figure}
In this section, the performance of the proposed method is studied in different cases including with and without data-augmentation. Moreover, we investigate the quality of the generated synthetic data by comparing them with the original data. We then apply the LIME method to determine the most important superpixels and regions of the input samples to provide explainability. Finally, comparisons with previous methods is presented in this section. 

To evaluate the classification ability of the proposed method, the following criteria are used \cite{bagherzadeh2022detection}:

\begin{equation}
\begin{array}{cc}
     Accuracy &=  \frac{TP+TN}{TP+TN+FP+FN}  \\
     Sensitivity& = \frac{TP}{TP+FN}\\
     Specificity &= \frac{TN}{TN+FP}\\
     \textit{F1-score} &= \frac{TP}{TP+0.5(FP+FN)}
\end{array}
    \label{eq:metrics}
\end{equation}
where $TP$, $TN$, $FP$, and $FN$ denote, true positive, true negative, false positive
and false negative elements (Positive class: patient with schizophrenia).

\subsection{Initial Classification} 
\label{sec:initial-classification}
In order to gain insight into the effectiveness of our proposed CNN architecture, it is compared to three other major successful CNN architectures, namely VGG-16, ResNet-50, and MobileNet-V2, to assess its effectiveness. These architectures have 65.0 million, 23.5 million, and, 2.3 million parameters respectively for 128x128 inputs. The VGG-16 architecture, despite having nearly 50 times more parameters than our model, demonstrates slightly lower accuracy on the test dataset and achieves 100\% accuracy on the training data, indicating relative overfitting. Consequently, we explored other renowned CNN architectures with fewer parameters, aiming to enhance the results over those achieved by VGG-16. However, as shown in Table \ref{tab:initialclf}, the outcomes were less favorable, suggesting that these models are not suitable for our classification task.
The proposed architecture outperforms MobileNet-V2 and ResNet-50 architectures by a significant margin, while reaching a slightly better accuracy than VGG-16. Table \ref{tab:initialclf} shows the result of the classification of two datasets using each of the three CNN architectures and the proposed CNN architecture over the test data. In this table, we have used the accuracy metric for the initial assessment and comparison of the four models to select the best for diagnosis. However, in subsequent sections, we present a comprehensive analysis of the selected model. For both datasets, 80\% of the instances were utilized as the training data, and the remaining 20\% were allocated for testing. Consequently, the number of training and testing samples for the 16-channel dataset is 806 and 202, respectively, while for the 19-channel dataset, the counts are 3315 for training and 829 for testing.
All models have been trained for 150 epochs (to convergence). After training, the initial model reaches a training accuracy of 100\%. The hyperparameters used are:
\begin{itemize}
\item \text{Optimizer: Adam}
\item \text{Learning rate: } $8 \times 10^{-5}$
\item \text{Batch size: 32}
\end{itemize}

The hyperparameters were selected through random search, insights from prior research, and commonly used values, customized to the specifics of our dataset and task. This methodology balanced systematic exploration with empirical guidance, further refined by iterative experimentation. The batch size, ranging from 16 to 64, was selected based on dataset size and typical benchmarks. The learning rate was determined through random search within commonly used ranges from previous studies. 

After completing the training process, our model demonstrates reduced overfitting compared to the other models listed in Table \ref{tab:initialclf}. Despite this improvement, it still exhibits a minor degree of overfitting. This issue can typically be addressed by expanding the training dataset, either through the addition of more data or by using data augmentation techniques to effectively increase dataset size.

\begin{table}
\centering
\caption{Average classification results of architectures on each  (test) dataset. Training accuracies are \textbf{100\%} for all}
\label{tab:initialclf}
\scriptsize
\begin{tabular}{lcccc}
\toprule
{Architecture}  & Proposed & VGG-16 & ResNet-50 & MobileNet-v2 \\
{Dataset}    & acc(\%) & acc(\%)  & acc(\%) & acc(\%) \\
\midrule
16-channel & 96.0 & 95.9  & 82.7 & 45.0 \\
19-channel & 99.6 & 96.4 & 97.7 & 52.5 \\
\bottomrule
\end{tabular}
\end{table}

\subsection{WGAN-GP and VAE for Data Synthesis} 
The initial classification results in Table \ref{tab:initialclf} indicate that the 19-channel dataset has performed excellently using the proposed architecture and achieved a near-perfect accuracy. Therefore, at this stage, we do not employ the generative augmentation method to augment the 19-channel dataset. However, the 16-channel dataset has achieved an accuracy of 96.0\% on the test data and a perfect accuracy on training data which indicates slight overfitting.

To address the overfitting observed in the 16-channel dataset, we employed generative augmentation techniques using Wasserstein GAN with gradient penalty (WGAN-GP) and variational autoencoder (VAE). By generating additional synthetic data, we aimed to improve the model's generalization capabilities and reduce overfitting. As mentioned in previous sections, separate WGANs and VAEs were used for each class.

The training configuration for each WGAN included the following hyperparameters:
\begin{itemize}
    \item Number of epochs: 2000
    \item Learning rate: $1 \times 10^{-4}$
    \item Discriminator training frequency: For every generator iteration, the discriminator was trained 3 times.
\end{itemize}

We selected these values by experimenting with the recommended values provided in the original WGAN-GP paper and the nearby values, while also accounting for the computational resources available. The remaining hyperparameters for WGAN-GP were chosen according to the suggestions made by the authors in the proposing paper \cite{gulrajani2017improved}. The network architectures were developed based on those used in the original paper, with necessary refinements and modifications to accommodate non-square input spectrograms. Figure \ref{fig:wgan} illustrates the schematics of the proposed rectangular WGAN-GP architecture for generating spectrograms. WGAN-GP yields high-fidelity samples with intricate details after approximately 500 epochs of training. Visual inspection further confirms the generation of a diverse range of synthetic samples. However, it is crucial to note that the original WGAN-GP paper highlights the need for extensive training, often exceeding 100K generator iterations, to achieve proper model convergence. this aspect makes the approach computationally expensive.  

The other generative network used is VAE and  The following training configurations were used for training each VAE:
\begin{itemize}
   
    \item Number of epochs: 6000
    \item Optimizer: Adam
    \item Learning rate: $8 \times10^{-5}$
    \item Latent representation vector size (z): 512
    \item Batch size: 32
   
\end{itemize}

The network architectures and all hyperparameters were selected considering the results of the initial classification in Section~\ref{sec:initial-classification}, the characteristics of our data and additional experimentation.
Every iteration for WGAN-GP takes approximately three times as long as VAE on the same GPU. Consequently, both models were trained for the same amount of time.

\begin{figure}[]
    \centering
    \includegraphics[width=\linewidth]{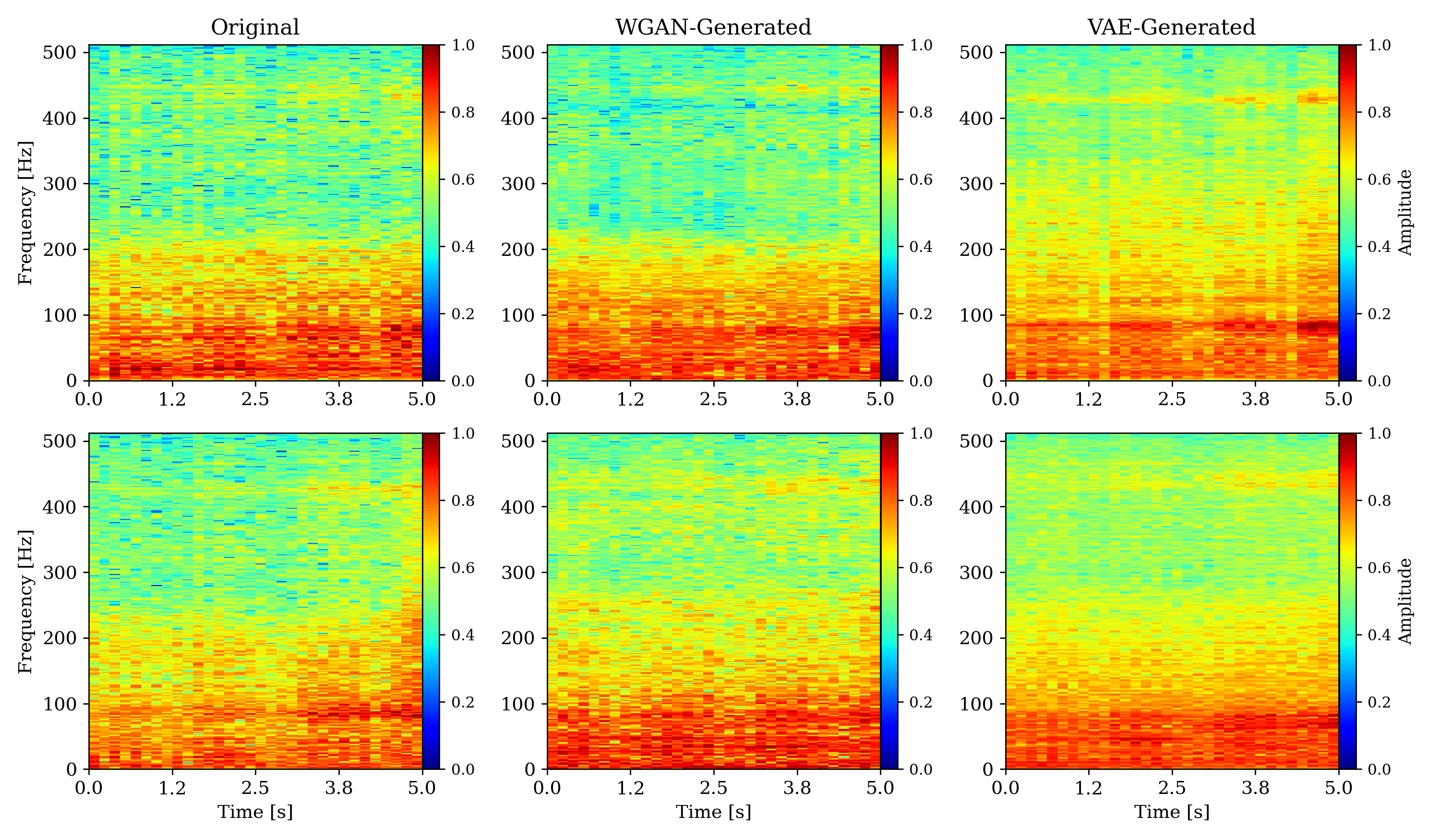}
    \caption{Visual comparison of the real data with data generated by WGAN and VAE, top row corresponds to normal samples and bottom row to schizophrenic}
    \label{fig:comp}
\end{figure}

\subsection{Evaluation of Synthetic Data}
One of the main challenges of using generative models is their evaluation since there is no ground truth to compare the generated samples with. Generative models are designed to generate samples that closely resemble the distribution of the original data. In the case of WGAN, its loss function is based on the Earth Mover's Distance (EMD), which serves as an estimate of the similarity between two distributions. The loss values resulting from this EMD representation guide the training of both the generator and discriminator in the WGAN framework. Therefore, one way to evaluate the WGAN model is to look at the loss values. Conversely, VAE loss values are a combination of the reconstruction loss and the Kullback-Leibler (KL) Divergence of the target distribution and the distribution of the learned latent representations which does not directly reflect the quality of the outputs and how close they are to the distribution of real data; but can provide a rough estimate for evaluation of the model.

To qualitatively analyse the performance of generative models a general approach is visual inspection. This method examines both the fidelity and diversity of the generated samples, ensuring high quality and resemblance to the original data while making sure that the generated data has properly captured the diversity of the original data. Proper visual inspection helps confirm that the model avoids mode collapse. In our study we aim at generating synthetic spectrograms and while visual inspection can give us a rough estimate of the performance of our generative models by looking at the high level features (e. g. brightness at certain frequency levels, etc.) in generated spectrograms, it is not a reliable method for analysis of the models robustness. Figure \ref{fig:comp} shows sample synthetic instances of each class for both models and based on the similarities between the original samples and synthetically generated samples we can conclude that both models are producing high-fidelity synthetic data.

To accurately assess our model’s performance, we must employ quantitative analysis to provide a reliable evaluation. We can not compare our generated images pixel by pixel to the original images, we can, however, compare and evaluate high-level features that are extracted by convolutional neural networks (CNN) or other feature extraction methods. Most of the proposed methods for the evaluation of generative models work based on this fundamental idea. 

To numerically evaluate our generative models, we have utilized the Train on Synthetic, Test on Real (TSTR) and Train on Real, Test on Synthetic (TRTS) metrics \cite{en13010130}. These metrics are introduced by Esteban et al. \cite{esteban2017real}. In the TSTR approach, a model is trained using a synthetic dataset and evaluated on a held-out set of real data, with accuracy being calculated. Conversely, in the TRTS approach, the model is trained on a dataset consisting entirely of real examples and then tested using a synthetic dataset. TSTR is considered the more ideal metric of the two because it evaluates the model's ability to generate both diverse and high-fidelity synthetic examples. In contrast, while TRTS measures how closely synthetic data resembles the original data, it may not reflect a lack of diversity and the occurrence of mode collapse in the model. Table \ref{tab:tstr-trts} shows the TSTR and TRTS average accuracies for VAE and WGAN-GP models. The training and test sets in this experiment consist of 900 and 200 samples respectively, with balanced representation from both classes and all models are trained to convergence using the learning rate of $8 \times 10^{-5}$, Adam optimizer and other 
hyperparameters similar to those detailed in Section~\ref{sec:initial-classification}. This experiments shows high TSTR and TRTS values for both models. However, it is evident that the VAE model outperforms WGAN-GP model by a significant margin. TSTR accuracy for the VAE model reaches 94.8\% which is slightly lower than the classification accuracies for the real training dataset tested on the real test data (shown in Table \ref{tab:initialclf}).

\begin{figure}[]
    \centering
    \includegraphics[width=1\linewidth]{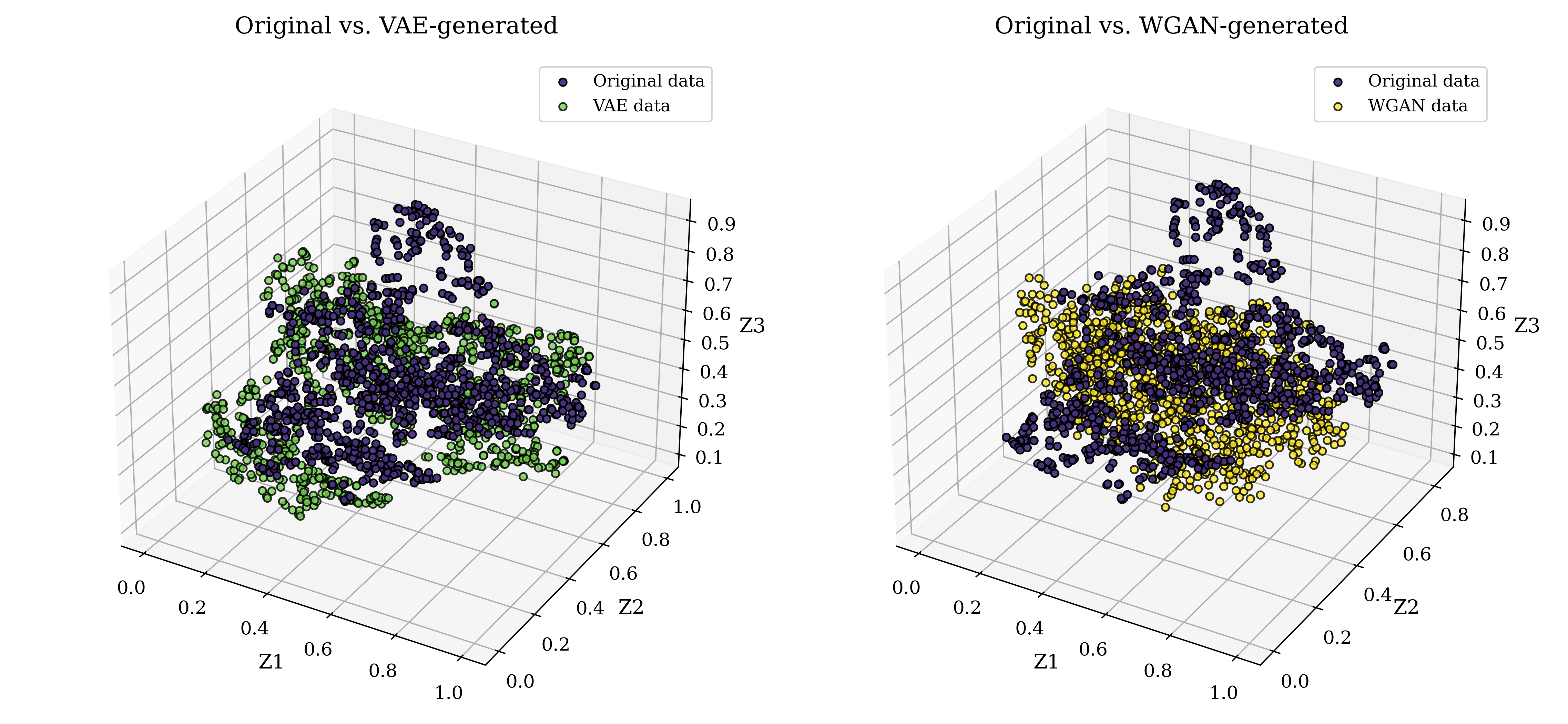}
    \caption{Visualization of the 3 principal components (Z1, Z2, Z3) of the synthetic datasets and the original data obtained using autoencoder and t-SNE}
    \label{fig:pca}
\end{figure}

Another very insightful approach for the evaluation of the synthetic data is to visually compare the distribution of them with the original data. We can not simply visualize image datasets due to high dimensionality. However, it is possible to reduce the dimensionality of the data down to its principal components and then visualize them. To reduce the dimensionality of our synthetic data we have generated 1008 (equal to the original dataset size) samples using each generative model and used a standard autoencoder to obtain latent representations in the shape of a vector of size 1024. Subsequently, using the t-SNE algorithm \cite{article} we reduce the dimensions of these vectors down to 3 to be able to visualize the data points. The result of this visualization is shown in Figure \ref{fig:pca}, which further confirms the results of Table \ref{tab:tstr-trts} since the VAE distribution more closely resembles the original data distribution.

\begin{table}
\centering
\caption{Train on Synthetic, Test on Real (TSTR) and Train on Real, Test on Synthetic (TRTS) accuracies for VAE and WGAN models. }
\label{tab:tstr-trts}
\footnotesize
\begin{tabular}{lcccccccccc}
\toprule
{metric}     & VAE (\%) & WGAN (\%) \\
\midrule
TSTR        & 94.8   & 80.5   \\
TRTS        & 98.1  & 80.5    \\
\bottomrule
\end{tabular}
\end{table}

\subsection{Augmented Classification}
This stage of our study includes augmenting the 16-channel dataset using the synthetic data generated by both VAE and WGAN-GP and evaluating the classification results. In order to correctly evaluate our augmentation method we have to first ensure that no synthetic data would be leaked into the test dataset. Therefore, we first separate 20\% of our dataset (which equals 202 instances) for testing before augmenting the training dataset. Datasets generated by each model require different learning rates in order to learn optimally. Therefore, we employed a learning rate of $1 \times 10^{-5}$ across all datasets in this experiment, as this value, the lowest obtained through random search, ensures consistency in our approach and we can cover all datasets with the same learning rate. In order to find the optimum number of synthetic datasets to be added to the original data, we have experimented with different values starting as low as 200 added synthetic data and incrementing 100 or 200 samples at every time. It is also important to mention that the original training dataset is slightly imbalanced; as a result, we add 30 more healthy (norm) samples to mitigate this small imbalance. Table \ref{tab:augclf} shows the results of this experiment in terms of accuracy and loss value. All classifiers are trained to convergence using our proposed CNN architecture.

\begin{table}
\centering
\caption{Accuracy \& loss of the model trained on augmented datasets of various added samples, over all-real test data with learning rate of $1e-5$ which is used only in this experiment.}
\footnotesize
\label{tab:augclf}
\begin{tabular}{lcccccccccc}
\toprule
{}              & \multicolumn{2}{c}{VAE} & \multicolumn{2}{c}{WGAN} \\
{norm, sch}     & acc(\%) & loss  & acc(\%) & loss \\
\midrule
+230, 200        &  96.5  & 0.158  &  92.6  & 0.165 \\
+330, 300        &  95.5  & 0.107  &  96.0  & 0.200 \\
+430, 400        &  97.5  & 0.075  &  96.0  & 0.233 \\
+630, 600        &  98.0  & 0.096  &  95.5  & 0.275 \\
\textbf{+730, 700}        &  \textbf{98.5}  & \textbf{0.016}  &  \textbf{96.5}  & \textbf{0.154} \\
+830, 800        &  98.0  & 0.068  &  92.1  & 0.397 \\
\bottomrule
\end{tabular}
\end{table}

\begin{figure}[]
    \centering
    \includegraphics[width=1\linewidth]{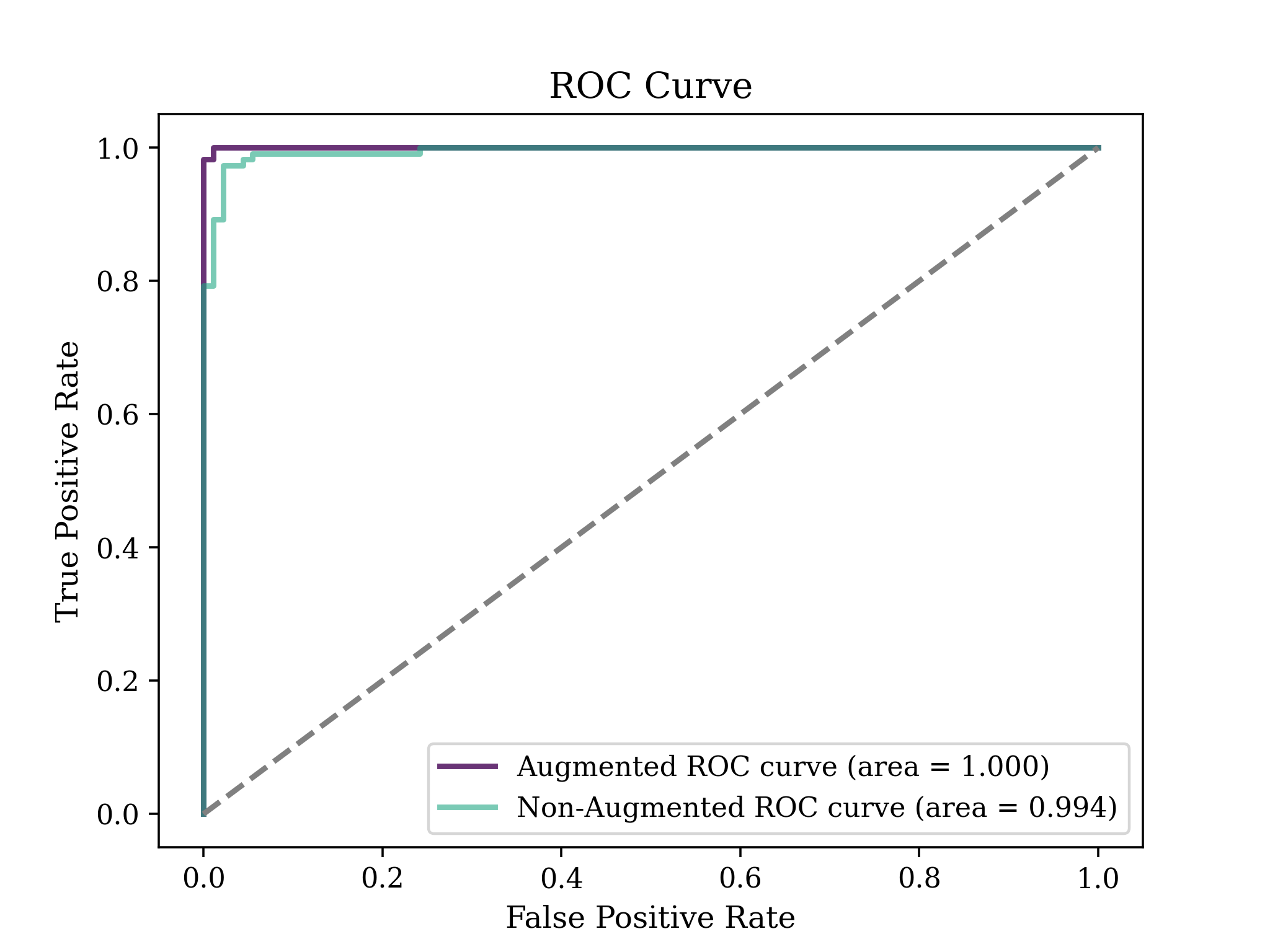}
    \caption{ROC (Receiver Operating Characteristic) curve of the "VAE-700" augmented model (left), and non-augmented model (right) with their respective Area Under the Curve (AUC)}
    \label{fig:rocs}
\end{figure}

This experiment makes it clear that the best generative model in this study for data augmentation is the VAE model and the optimal number of instances to be added are 730 and 700 for norm (healthy) and sch (schizophrenic) classes respectively. Subsequently, we carried out a separate experiment with the superior augmented model and dataset and the non-augmented dataset to conduct a more insightful comparison and analysis of the augmented and non-augmented classifications. For this experiment, we trained a classifier with our proposed CNN model on both datasets (augmented and non-augmented) with the same training configurations as before. We call the augmented dataset "VAE-700" since it uses VAE generative model and 700, 700 + 30 added instances for each class. The augmented model reached an accuracy of 99.0\% compared to 96.0\% accuracy for the non-augmented model while also reaching convergence in fewer epochs and having smoother training curves. This indicates that the generative augmentation using VAE significantly improves the performance of the baseline schizophrenia diagnosis model. Table \ref{tab:final_eval} compares the two models using several evaluation metrics which gives us a better insight into the performance of both models and Figure \ref{fig:rocs} compares the ROC curves and Area Under the Curve (AUC) for both models. These comparisons using several metrics eliminate any uncertainty regarding the improvement achieved using the VAE-augmented dataset.

\begin{table}
\centering
\caption{The comparison of augmented and non-augmented models with several metrics over the \textbf{test} dataset, the training accuracy is \textbf{100\%} for both models after convergence}
\label{tab:final_eval}
\footnotesize
\begin{tabular}{lcccccccccc}
\toprule
{}              & {VAE-700} & {Non-augmented} \\
{metric}        &           &           \\
\midrule
Accuracy(\%)    &  99.0    &  96.0  \\
F1-score        &  0.990   &  0.960 \\
Sensitivity     &  0.992   &  0.958 \\
Specificity     &  0.986   &  0.962 \\

\bottomrule
\end{tabular}
\end{table}

\begin{table*}[t]
\centering
\caption{Comparison of our study results with state-of-the-art studies in deep learning-based diagnosis of schizophrenia on dataset with 16 channels (45 SZ patients and 39 healthy subjects). Our results are represented as the mean accuracy of 5 repetitions of the experiment, and the standard deviations are negligible.}
% Define a new centered version of the p{width} column type
\newcolumntype{C}[1]{>{\centering\arraybackslash}p{#1}}

\label{tab:compar1}
\footnotesize
\begin{tabular}{C{3cm}C{3.5cm}C{3cm}C{3.0cm}C{1.5cm}}
\toprule
{Ref}  & {EEG Dataset} & {Feature Extraction} & {Architecture and Methods} & {Accuracy}\\
{}         & {}            & {}                   & {}           & {}     \\
\midrule

Aslan and Akin \cite{aslan2020automatic} & 45 SZ patients, 39 healthy (16 channels)   
& Time-Frequency, Spectrogram (STFT) & CNN, VGG-16 & 95.0\% \\
\\
Phange et al. \cite{8836535} & 45 SZ patients, 39 healthy (16 channels) & Time-Frequency connectivity, Complex network measures & Multi-domain CNN (1D + 2D CNN) & 91.69\%\\
\\
Naira et al. \cite{Naira2019} & 45 SZ patients, 39 healthy (16 channels) & Pearson Correlation Coefficients (PCC) matrix & CNN & 90.0\%\\
\\

Sobahi et al. \cite{sobahi2022new} & 45 SZ patients, 39 healthy (16 channels) & 1D local binary pattern (LBP) & ELM-based autoencoders augmentation, Transfer learning & 97.7\%\\
\\

\textbf{Our study (Generative Augmentation on 16-channel dataset)} & \textbf{45 SZ patients, 39 healthy (16 channels)} & \textbf{Time-Frequency, Spectrogram arrays (STFT)} & \textbf{CNN, VAE generative network} & \textbf{99.0\%} \\
\\

\textbf{Our study (16-channel dataset without Augmentation)} & \textbf{45 SZ patients, 39 healthy (16 channels)} & \textbf{Time-Frequency, Spectrogram arrays (STFT)} & \textbf{CNN} & \textbf{96.0\%} \\
\\
\bottomrule
\end{tabular}
\end{table*}

\begin{table*}[t]
\centering
\caption{Comparison of our study results with state-of-the-art studies in deep learning-based diagnosis of schizophrenia on dataset with 19 channels (14 SZ patients and 14 healthy subjects). Our results are represented as the mean accuracy of 5 repetitions of the experiment, and the standard deviations are negligible.}
% Define a new centered version of the p{width} column type
\newcolumntype{C}[1]{>{\centering\arraybackslash}p{#1}}

\label{tab:compar2}
\footnotesize
\begin{tabular}{C{3cm}C{3.5cm}C{3cm}C{3cm}C{1.5cm}}
\toprule
{Ref}  & {EEG Dataset} & {Feature Extraction} & {Architecture and Methods} & {Accuracy}\\
{}         & {}            & {}                   & {}           & {}     \\
\midrule
Shalbaf et al. \cite{shalbaf2020transfer} & 14 SZ patients, 14 healthy (19 channels)  
& Time-Frequency, Scalogram (CWT) & Transfer learning, CNN, SVM & $98.60\% \pm $2.29 \\
\\
Sahu et al. \cite{sahu2023scz} & 14 SZ patients, 14 healthy (19 channels)   
& Time-Frequency, Scalogram (CWT) & Depth-wise Separable Convolution, Attention & $99.0\% \pm $0.25 \\
\\
Aslan and Akin \cite{aslan2020automatic} & 14 SZ patients, 14 healthy (19 channels)   
& Time-Frequency, Spectrogram (STFT) & CNN, VGG-16 & 97.0\% \\
\\

Bagherzadeh et al. \cite{bagherzadeh2022detection} & 14 SZ patients, 14 healthy (19 channels) & Transfer entropy (TE) connectivity & Ensemble of hybrid pre-trained CNNs-LSTM
models & 99.9\%\\
\\
\textbf{Our study (19-channel dataset without Augmentation)} & \textbf{14 SZ patients, 14 healthy (19 channels)} & \textbf{Time-Frequency, Spectrogram arrays (STFT)} & \textbf{CNN} & \textbf{99.6\%} \\
\\

\bottomrule
\end{tabular}
\end{table*}

\iffalse
\begin{figure}[]
    \centering
    \includegraphics[width=\linewidth]{simple_curves.png}
    \caption{Training curves of the non-augmented model after 100 epochs of training}
    \label{fig:simple}
\end{figure}

\begin{figure}[]
    \centering
    \includegraphics[width=\linewidth]{vae700_curves.png}
    \caption{Training curves of the "VAE-700" augmented model after 100 epochs of training}
    \label{fig:vae700}
\end{figure}
\fi

\subsection{Explainablity Analysis} 
\begin{figure}[t]
    \centering
    \includegraphics[width=3.3 in]{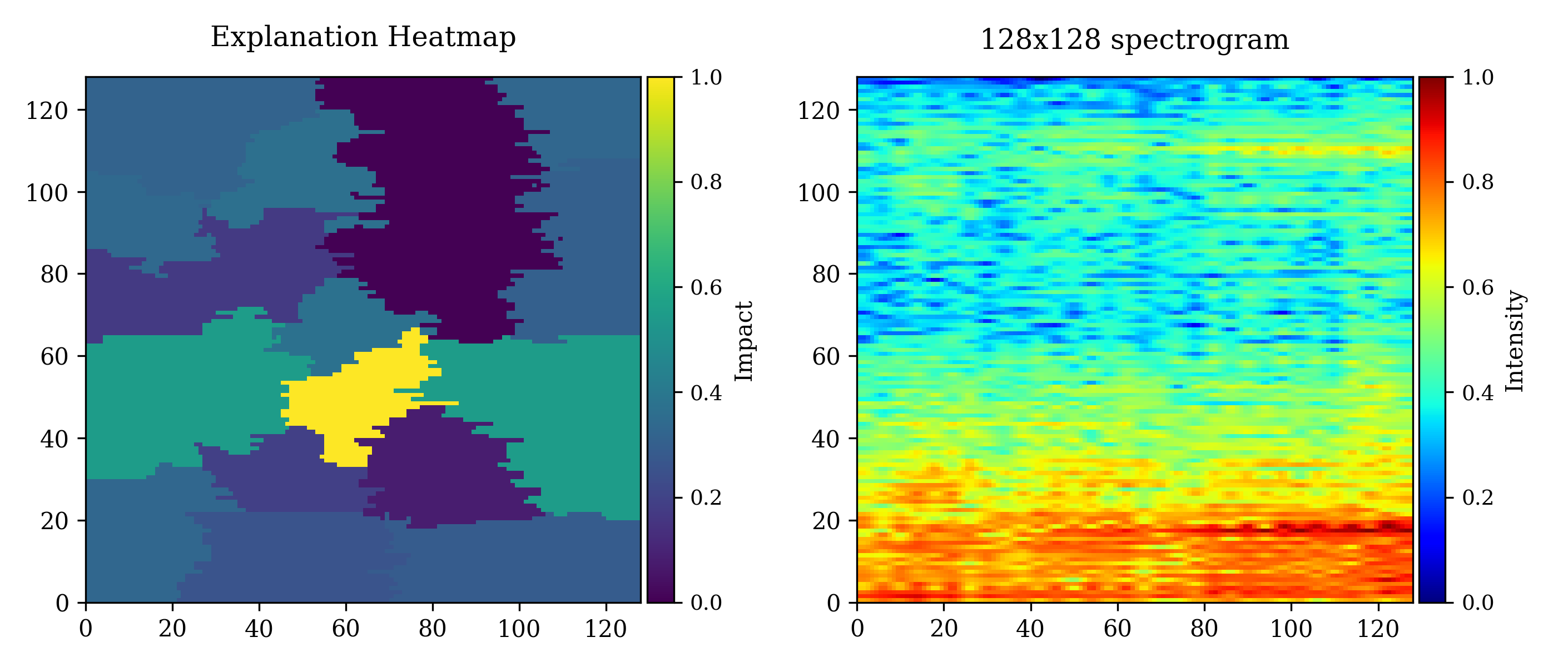}
    \caption{Explanations heatmaps for resized spectrogram image of a schizophrenic (sch) subject, indicating the impact of each superpixel on the prediction}
    \label{fig:xai-sch}
\end{figure}

\begin{figure}[t]
    \centering
    \includegraphics[width=3.3in]{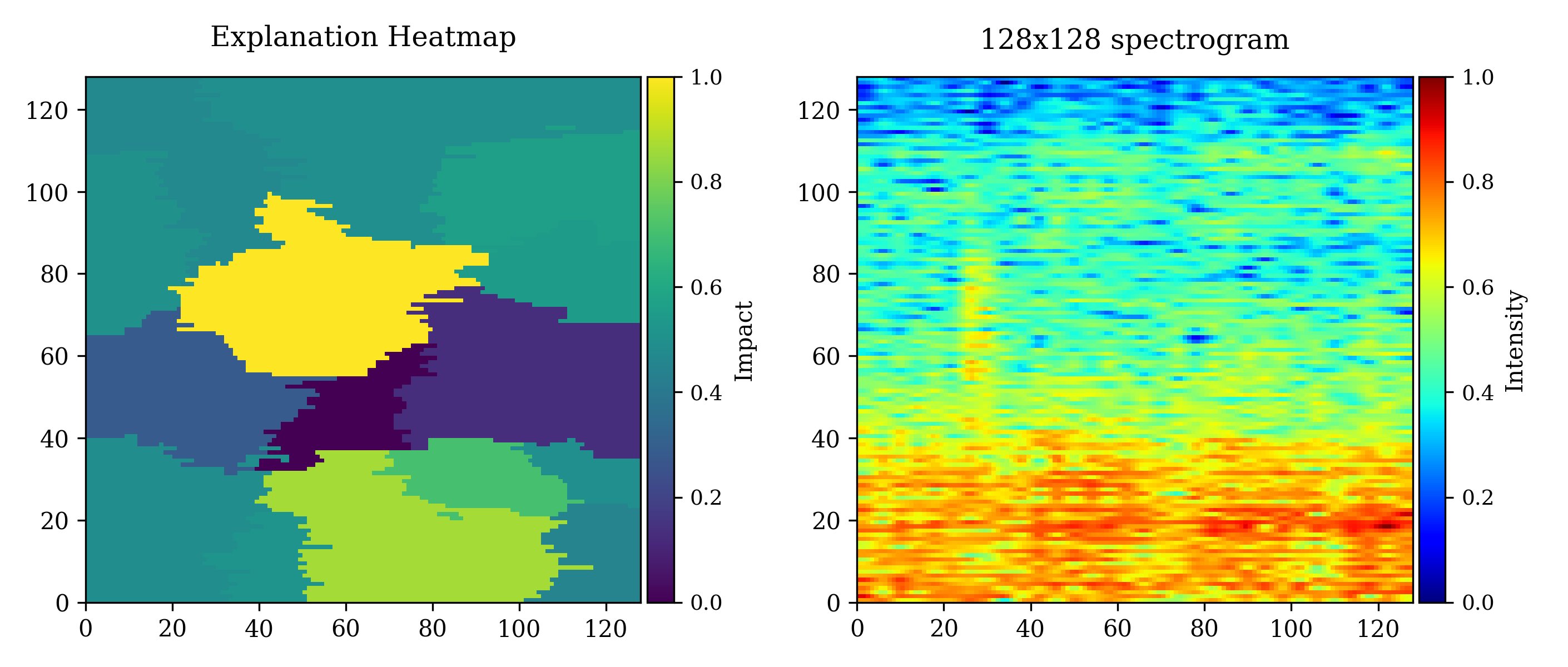}
    \caption{Explanations heatmaps for resized spectrogram image of a normal (norm) subject, indicating the impact of each superpixel on the prediction }
    \label{fig:xai-norm}
\end{figure}

\begin{figure}[t]
    \centering
    \includegraphics[width=3.3in]{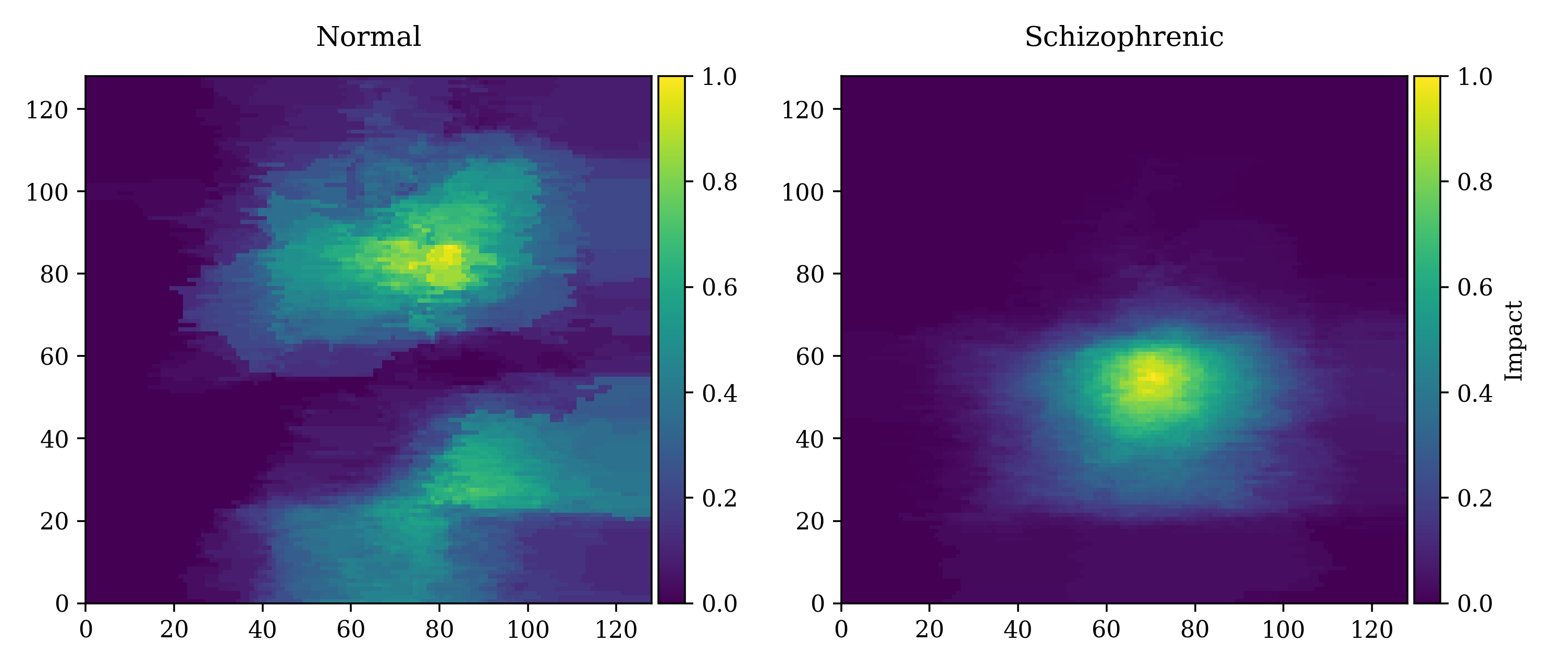}
    \caption{Average of the most impactful superpixel region for normal (left) and schizophrenic (right) spectrograms explanation heatmaps for the real data}
    \label{fig:acc-real-heatmap}
\end{figure}

\begin{figure}[t]
    \centering
    \includegraphics[width=3.3in]{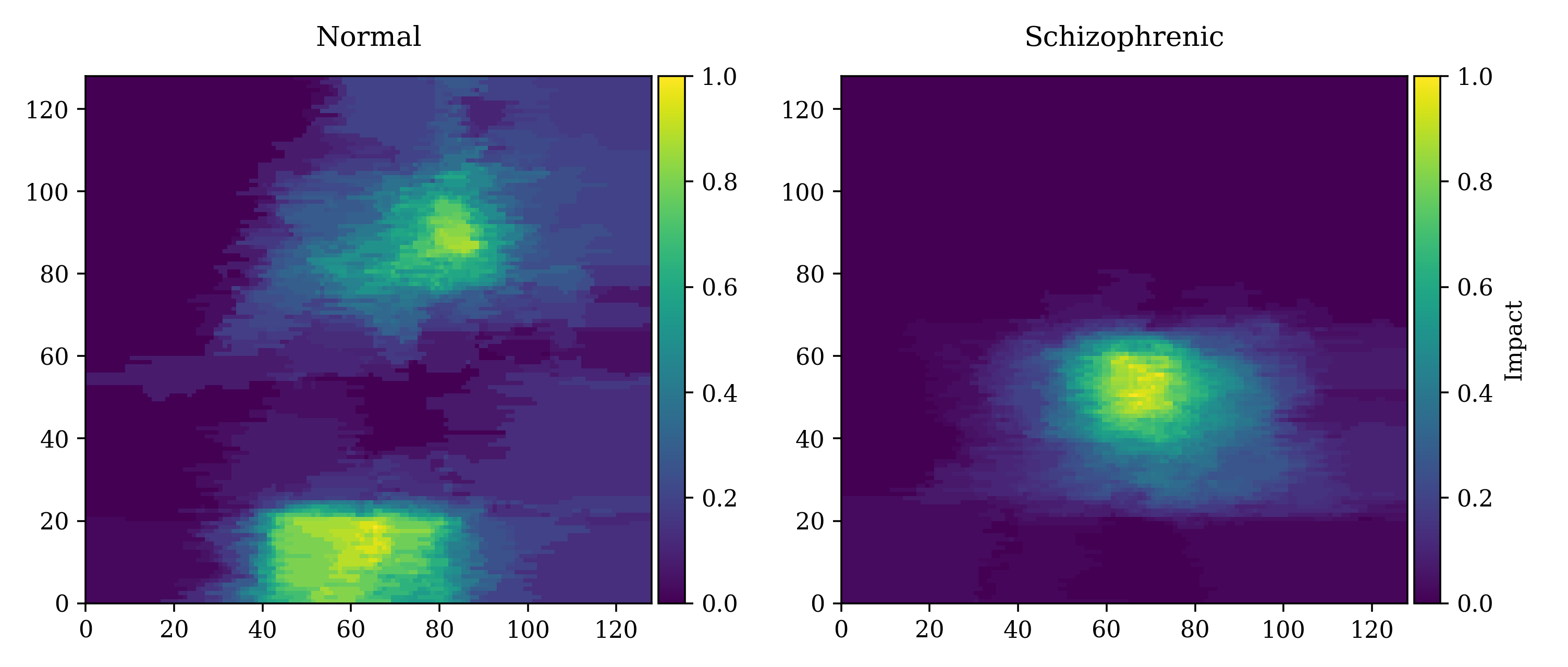}
    \caption{Average of most impactful superpixel region for normal (left) and schizophrenic (right) spectrograms explanation heatmaps for the synthetic data }
    \label{fig:acc-fake-heatmap}
\end{figure}

We apply the LIME algorithm to the trained classifier. The explanations provided by LIME are in the form of segmented superpixel regions, where the intensity level of each region indicates its impact on the classification decision for the respective class. Figures \ref{fig:xai-sch} and \ref{fig:xai-norm} show the explanation heatmaps for a sample schizophrenic and normal spectrogram, respectively. These heatmaps highlight different super-pixels in the spectrograms with varying levels of brightness. The brighter colors indicate higher significance in contribution. We extracted these heatmaps for all instances in the test dataset to identify consistent patterns.To facilitate intuitive analysis of these explanations, we retained the superpixel region with the maximum value and zeroed out the rest of the heatmap for each instance. We then calculated the average of these heatmaps. This method can help us verify if a consistent pattern exists in the explanation results of each class. Figure \ref{fig:acc-real-heatmap} illustrates the resulting array, displaying these average heatmaps. This figure clearly demonstrate the presence of a pattern in the explanations. For spectrograms belonging to schizophrenic patients, the regions of significance are distinctly different from those in normal spectrograms. These regions correspond to different frequency levels in the original signals. For the schizophrenic instances, the most significant region is between pixels 40 and 60 on the vertical axis, corresponding to frequency levels of around 160 to 240 Hz in the unresized spectrograms. In contrast, for normal instances, the most significant region is between pixels 70 and 90, which correspond to frequency levels of around 280 to 360 Hz and some areas below the 160 Hz level are also highlighted. The same procedure is performed for the synthetic instances generated by the VAE and the same patterns were present as shown in the Figure \ref{fig:acc-fake-heatmap}. Additionally, the consistency between the results of the explainer algorithm for real and synthetic instances further proves the robustness of our trained generative model and the quality of the generated instances.

\subsection{Comparison with Related Studies}
In this section, we compare our study results with related studies focused on the diagnosis of schizophrenia using deep learning methods. In Table \ref{tab:compar1} and Table \ref{tab:compar2} we provided a comprehensive list of the related works which have utilized deep learning methods for diagnosis of schizophrenia. These studies have used different network architectures and feature extraction methods. From the accuracy column of this table, it is evident that our method has outperformed all other studies which have used a similar dataset by a relatively large margin. Some of the studies on this list have used the 19-channel dataset which we have also used in this study. This dataset creates significantly more data instances due to its longer recording length, therefore, studies that have used this dataset achieved better accuracies compare to the 16-channel dataset. We chose not to augment the second (19-channel) dataset in this study since it achieved an impressive accuracy of 99.6\% using our proposed CNN architecture without the augmentation.  Reference \cite{bagherzadeh2022detection} is the only study among the studies in the list that has classified schizophrenia and healthy subjects slightly more accurately than our method on the 19-channel dataset. However, it is important to consider that the 99.6\% accuracy achieved by our method on the second dataset has been achieved without the generative data augmentation.
\section{Conclusion}
In this study, our goal was to present a deep learning-based approach in order to facilitate the diagnosis of schizophrenia and aid its early detection. Among all machine learning paradigms, supervised learning is the one that is commonly used for medical diagnosis.  

However, supervised learning, especially if used in deep learning, requires large datasets in order to be effective. In many medical applications, we can not easily access more data. Therefore, in this study, we have explored a data augmentation method using generative models in order to improve our classification result for the diagnosis of schizophrenia. Two approaches were employed, Wasserstein GAN with Gradient Penalty (WGAN-GP) and Variational autoencoder (VAE). It was concluded that VAE generated more similar samples to the original dataset and improved the accuracy of the classification more than WGAN-GP trained for a similar duration of time. However, it is important to consider that WGAN-GP can generate highly-detailed synthetic samples and when trained longer, could potentially prove to be the better generative approach between the two methods. 

We were also able to outperform the previous studies conducted on the same datasets by a significant margin. The classifier also exhibited very impressive results on original test data when trained only on the synthetic dataset. The results of this study in terms of data augmentation were promising and indicated that generative data augmentation could be very effective in the context of medical diagnosis of mental disorder specially for EEG data modality. 

Finally, we explained our model using the LIME algorithm. This approach can significantly help build trust in AI systems for medical diagnosis and assist specialists in identifying key indicators for disease diagnosis. Explainability is a crucial element in using deep learning for medical diagnosis, and every reliable study must include it to ensure the model is functioning correctly and to assure potential users of its reliability.

In future works, we can train WGAN-GP for more iterations to outperform our current VAE model and generate better-quality synthetic instances or potentially combine WGAN and VAE to make use of the advantages of both models. We can also employ our method in the diagnosis of other diseases based on EEG or ECG signals to improve their accuracy. Combining different modalities including fMRI with EEG could be another future study.
\bibliographystyle{ieeetr}
\bibliography{mybibfile} 
%\vspace{2.25cm}
%\newpage
\vspace{-4cm}
\begin{IEEEbiography}
[{\includegraphics[width=1in,height=1.25in,clip,keepaspectratio]{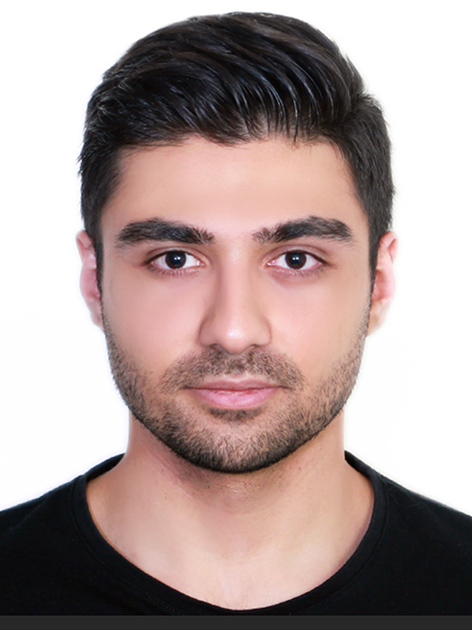}}]{Mehrshad Saadatinia}
  earned his B.Sc. degree in Computer Engineering from Shahid Beheshti University, Tehran, Iran, in 2023. He has briefly worked as a Research Assistant at Shahid Beheshti University Natural Language Processing (NLP) and \textit{Robotics \& Intelligent Autonomous Agents (RoIAA)} Laboratories. He was the recipient of the best B.Sc. Thesis Award in the Computer Engineering Department for his work on schizophrenia diagnosis. His primary research interests lie in the fields of Deep Learning and Computer Vision, with a specialized focus on their applications in Medical Diagnosis and Precision Medicine. 
\end{IEEEbiography}
%\newpage
\vspace{-4cm}
\begin{IEEEbiography}
    [{\includegraphics[width=1in,height=1.25in,clip,keepaspectratio]{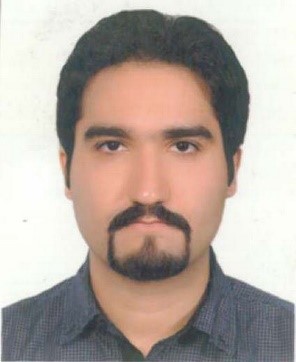}}]{Armin Salimi-Badr (Senior Member)}
 received the B.Sc., M.Sc. and PhD degrees in Computer Engineering, all from Amirkabir University of Technology, Tehran, Iran in 2010, 2012, and 2018 respectively. He also obtained a PhD degree in Neuroscience from University of Burgundy, Dijon, France in 2019, where he was researching on presenting a computational model of brain motor control in the Laboratory 1093 CAPS (Cognition, Action, et Plasticité Sensorimotrice) of the Institut National de la Santé et de la Recherche Médicale (INSERM). He was a Postdoctoral Research Fellow at Biocomputing lab of Amirkabir University of Technology from October 2019 to September 2020.
Currently, he is an Assistant Professor at Faculty of Computer Science and Engineering of Shahid Beheshti University, Tehran, Iran and also the Head of \textit{Artificial Intelligence\&Robotics\&Cognitive Computing} group in this faculty. He is also the founder and Chair of \textit{Robotics \& Intelligent Autonomous Agents (RoIAA)} Lab in Shahid Beheshti University. He is currently the Chair of Professional Activities Committee and a Board Member of Computer Society of IEEE Iran Section. His research interests include Computational Intelligence, Computational Neuroscience, and Robotics.
\end{IEEEbiography}
\EOD

\end{document}